\pdfoutput=1
\documentclass{article}
\PassOptionsToPackage{numbers,sort&compress}{natbib}
\usepackage{paper,times}
\usepackage{amsmath,amssymb}
\usepackage{xcolor}
\usepackage{graphicx}
\usepackage{booktabs}
\usepackage{makecell}
\usepackage{caption}
\usepackage{placeins}
\usepackage{float}
\usepackage{microtype}
\usepackage{url}
\usepackage{hyperref}

\definecolor{NvidiaGreen}{RGB}{118,185,0}
\hypersetup{
    colorlinks=true,
    citecolor=NvidiaGreen,
    linkcolor=NvidiaGreen,
    urlcolor=NvidiaGreen,
    pdftitle={Should the Action Expert Train the VLM? Action-Token Supervision and Gradient Routing in Mixture-of-Transformer VLAs},
    pdfauthor={Asher James Hancock, Boyi Li, Yuxiao Chen, Shuhan Tan, Wenhao Ding, Thomas Tian, Yan Wang, Boris Ivanovic, Marco Pavone}
}

\title{Should the Action Expert Train the VLM?\\Action-Token Supervision and Gradient Routing in Mixture-of-Transformer VLAs}
\author{\parbox[t]{\textwidth}{\normalfont\raggedright
Asher James Hancock\textsuperscript{1,2,*},
Boyi Li\textsuperscript{1,\textdagger},
Yuxiao Chen\textsuperscript{1},
Shuhan Tan\textsuperscript{1},
Wenhao Ding\textsuperscript{1},
Thomas Tian\textsuperscript{1},
Yan Wang\textsuperscript{1},
Boris Ivanovic\textsuperscript{1},
Marco Pavone\textsuperscript{1,3}\\[4pt]
\textsuperscript{1}NVIDIA \quad
\textsuperscript{2}Princeton University \quad
\textsuperscript{3}Stanford University\\
\textsuperscript{*}Work done during internship at NVIDIA \quad
\textsuperscript{\textdagger}Corresponding author\\[4pt]
\url{https://mot-vla-gradient-routing.github.io/}
}}

\begin{document}
\maketitle

\begin{abstract}
Vision--language--action (VLA) models increasingly pair a pre-trained vision--language model (VLM) with a flow-matching expert that conditions its action predictions on the VLM's key--value representations. Those representations must carry information useful for control, yet the VLM was never trained for the action expert's flow-matching objective. Two signals can address this: gradients backpropagated from the expert's loss (\emph{coupling}), or VLM supervision to predict discrete tokens encoding the same target actions. Prominent training recipes forgo the first, blocking expert-to-VLM gradients (\emph{insulation}) to protect the VLM's existing capabilities, while also supervising it on vision--language data and action tokens. Because vision--language co-training also aims to preserve those capabilities, the additional protection provided by insulation is unclear. At the same time, insulation removes a direct signal about which features continuous action prediction requires. On the other hand, action-token supervision offers an alternative way to learn action-relevant representations, which may explain why insulation succeeds in practice. We disentangle these factors in autonomous driving and manipulation by independently varying action-token supervision and whether the expert's loss updates the VLM. In open-loop driving, insulation is competitive with coupling when action tokens are supervised; without them, coupling reduces trajectory-prediction error by roughly 25\%. When both components are trained jointly from the outset, coupling increases closed-loop driving route progress by approximately 17\% with action-token supervision and 40\% without it, and similarly improves robot manipulation performance under both settings. These findings show that \textbf{the value of gradient insulation depends on the VLM's action supervision and the evaluation setting}, challenging insulation as a default VLA training recipe.

\end{abstract}

\begin{figure}[t]
\centering
        
        \includegraphics[width=\linewidth]{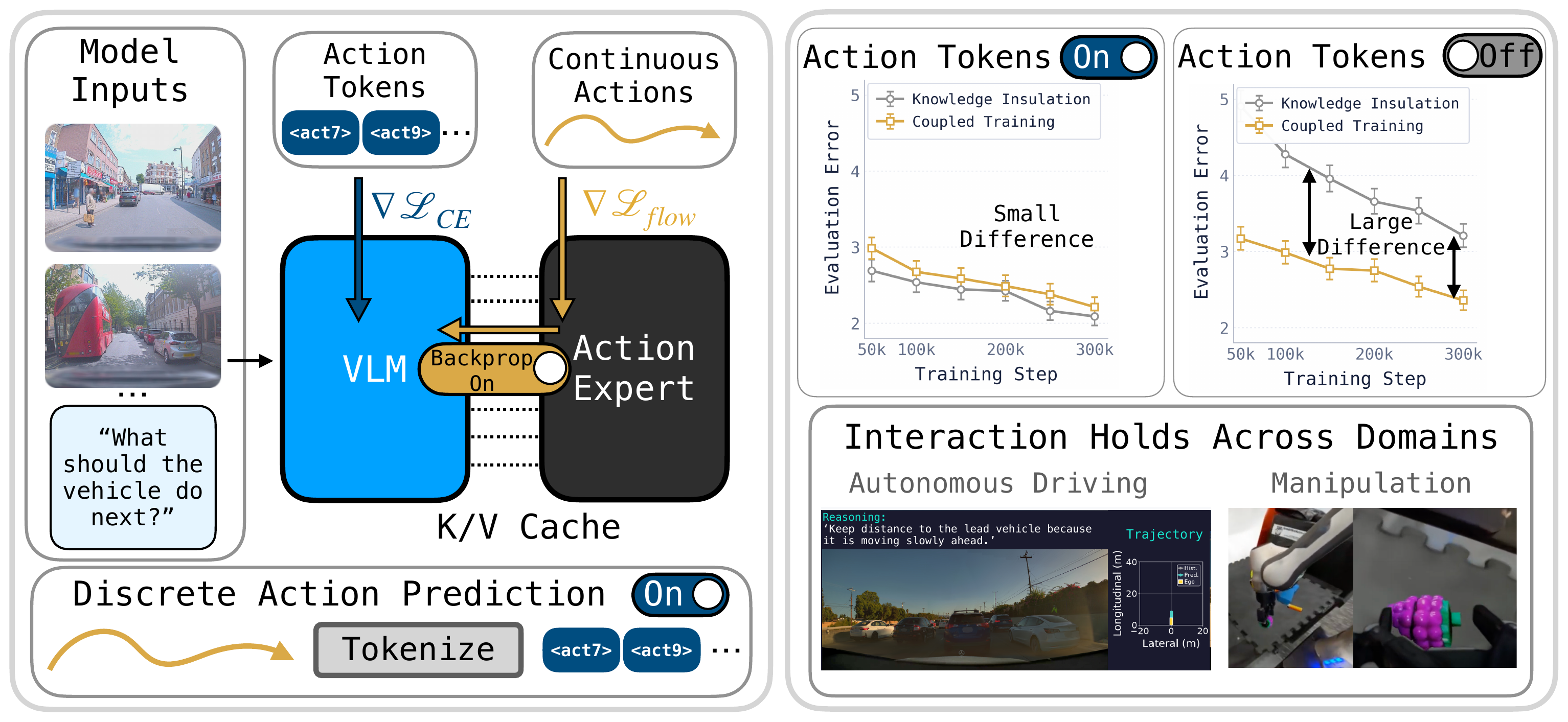}
        \caption{\textbf{Action-token supervision shapes the effect of expert-to-VLM
gradient flow.} \textbf{Left:} the VLM is supervised to predict discrete action
tokens while a flow-matching expert decodes
continuous actions from the VLM's K/V cache. We independently vary whether
action tokens are supervised and whether the expert's gradient
($\nabla\mathcal{L}_{\mathrm{flow}}$) updates the VLM. \textbf{Right:} open-loop
driving evaluation error during training. With action-token supervision, insulated and
coupled training differ little; without it, coupling is substantially better. We test this interaction in closed-loop autonomous driving and robot manipulation.}
        \label{fig:fig1}
\end{figure}

\section{Introduction}
\label{sec:introduction}

Pretrained vision--language models (VLMs) bring broad perceptual, semantic, and reasoning capabilities to embodied control. Robotic control, however, requires precise continuous outputs at frequencies poorly suited to autoregressive decoding with large VLMs. Many recent vision--language--action (VLA) models therefore use Mixture-of-Transformers-style architectures \citep{liang2024mixture}, pairing an autoregressive VLM with a smaller diffusion or flow-matching action expert \citep{black2024pi_0}. The VLM processes visual observations and language instructions, while the expert efficiently generates continuous action chunks conditioned on the VLM's per-layer key--value (K/V) representations \citep{intelligence2025pi_05}.

This architecture leaves open a consequential training decision: should the continuous-action loss train only the expert, or also the VLM that produces its conditioning K/V cache? Allowing expert gradients to update the VLM (\emph{coupling}) enables co-adaptation and can teach the backbone which information its representations must preserve for precise control. Yet it also exposes pretrained visual and linguistic features to gradients from a newly introduced, often randomly initialized expert, which may degrade the VLM's semantic and world understanding.

Popular approaches therefore block the action expert's gradients from entering the VLM (\emph{insulation}), either through stop-gradient or frozen-backbone training \citep{intelligence2025pi_05,driess2026knowledge}. However, full-parameter fine-tuning on a narrow robot training distribution can degrade pretrained capabilities regardless of gradient routing \citep{hancock2025actionslanguagefinetuningvlms}. When vision--language co-training is also used to mitigate forgetting, it is unclear whether insulation provides additional protection.

Insulation is also rarely evaluated in isolation. Blocking expert gradients removes one source of action-relevant learning signal, but such systems typically supply another: they train the VLM to predict discrete action tokens \citep{pertsch2025fast}, which directly supervise the behavior the expert models. Co-training on general vision--question-answering (VQA) and chain-of-thought examples may preserve the VLM's semantic capabilities, but it does not explicitly encourage representations relevant to control. In the absence of action-token supervision, insulation may therefore eliminate the only objective that directly adapts the VLM for trajectory prediction. Crucially, defining an analogous discrete proxy for continuous prediction targets is a domain-dependent design choice whose effectiveness must be established empirically. It is therefore unclear whether the reported benefits of insulation arise from gradient blocking itself or depend on this auxiliary supervision.

Prior work has not isolated this interaction. Knowledge Insulation reports benefits from insulated training while supervising the VLM with FAST action tokens and vision--language objectives, but does not evaluate insulation without FAST supervision \citep{driess2026knowledge,pertsch2025fast}. In contrast, Wall-OSS-0.5 reports stronger action performance from coupled training even when discrete-action and multimodal objectives are retained \citep{yu2026walloss05technicalreport}. Because these systems also differ in training recipes, initialization, data, and optimization, their seemingly opposing results cannot isolate the contributions of action-token supervision and expert-to-VLM gradient flow in downstream control or determine how the two mechanisms interact.

\textbf{Contributions.}
In this work, we study how a VLA's representations are made action-relevant, and whether the two available signals---the expert's gradient and discrete action-token supervision---substitute for one another. We cross action-token supervision with expert-to-VLM gradient flow in autonomous driving and robot manipulation, under both joint training and training schedules in which one component is adapted first. Within each training schedule and action-token setting, routing comparisons hold data, architecture, optimization, and semantic co-training fixed. In open-loop driving evaluations, the two routing choices are nearly indistinguishable with action-token supervision; without it, coupling reduces trajectory-prediction error by roughly 25\%. This pattern holds across training schedules and in both in- and out-of-distribution evaluations. Under joint training, coupling improves closed-loop performance regardless of supervision: driving route progress increases by approximately 17\% with action tokens and 40\% without them, while manipulation task progress improves under both settings. Together, these results identify action-token supervision as a factor previously entangled with gradient routing and show that \textbf{VLM insulation is a conditional choice, not a default training recipe for training VLAs}.

\section{Related Work}
\label{sec:related_work}

\textbf{Vision--language--action architectures.}
Early VLA policies incorporated control into pretrained vision--language models (VLMs) by predicting discretized actions autoregressively \citep{brohan2023rt2visionlanguageactionmodelstransfer,kim2024openvlaopensourcevisionlanguageactionmodel}. Many recent systems instead pair a VLM with a specialized continuous-action module. $\pi_0$ introduced a flow-matching expert conditioned on the VLM's per-layer key--value (K/V) cache, an architecture retained or adapted by $\pi_{0.5}$, Wall-OSS-0.5, and MolmoAct2 \citep{black2024pi_0,intelligence2025pi_05,yu2026walloss05technicalreport,fang2026molmoact2actionreasoningmodels}. SmolVLA, GR00T~N1, Qwen-RobotManip, CogACT, and DexVLA use related flow- or diffusion-based action decoders with different interfaces to the VLM \citep{shukor2025smolvlavisionlanguageactionmodelaffordable,nvidia2025gr00tn1openfoundation,yuan2026qwenrobotmaniptechnicalreportalignment,li2024cogactfoundationalvisionlanguageactionmodel,wen2025dexvlavisionlanguagemodelplugin}. These systems instantiate a broader Mixture-of-Transformers pattern in which modality-specific computation communicates through attention \citep{liang2024mixture}.

Many autonomous-driving VLAs follow similar designs. EMMA, OpenDriveVLA, and AutoVLA generate future trajectories autoregressively as textual coordinates or discrete tokens, whereas Alpamayo-R1 combines autoregressive physical reasoning and trajectory prediction with a continuous action expert \citep{hwang2025emmaendtoendmultimodalmodel,zhou2025opendrivevlaendtoendautonomousdriving,zhou2025autovlavisionlanguageactionmodelendtoend,wang2025alpamayo}. Related architectures combine autoregressive reasoning with continuous generation for images, video, and other physical-AI modalities \citep{zhou2024transfusionpredicttokendiffuse,nvidia2026cosmos3omnimodalworld}. Despite their growing adoption, how continuous experts shape the autoregressive representations through the K/V cache they condition on remains poorly understood. We study this question in a controlled setting.

\textbf{Action representation and backbone supervision.}
Action representation determines both how a policy is executed and how control supervision reaches the VLM. Discrete action tokens train the backbone through next-token prediction, while diffusion and flow matching directly model continuous, potentially multimodal action distributions \citep{chi2024diffusionpolicyvisuomotorpolicy,black2024pi_0}. FAST compresses action chunks into discrete tokens using a discrete cosine transform, OAT learns compact action tokens whose prefixes decode into progressively refined action chunks, OpenVLA-OFT predicts normalized continuous action chunks in parallel through an MLP head, and HybridVLA combines autoregressive and diffusion-based action prediction \citep{pertsch2025fast,kim2025finetuningvisionlanguageactionmodelsoptimizing,liu2026orderedactiontokensvisuomotor, liu2025hybridvlacollaborativediffusionautoregression}.

\begin{table}[!htbp]
\centering
\caption{Driving and manipulation VLAs that block action-expert gradients from the VLM through freezing or knowledge insulation \citep{driess2026knowledge}. Checkmarks indicate VQA or discrete action-token supervision.}
\label{tab:insulated_vla_recipes}
\footnotesize
\setlength{\tabcolsep}{4pt}
\begin{tabular}{@{}p{0.48\linewidth}cc@{}}
\toprule
Policy & \makecell{VQA\\Predicted} & \makecell{Discrete Actions\\Predicted} \\
\midrule
\multicolumn{3}{@{}l}{\textbf{Autonomous Driving}} \\
Alpamayo-R1 \citep{wang2025alpamayo} & $\checkmark$ & $\checkmark$ \\
Qwen-Drive-1.0 \citep{zhou2026qwendrive10initialstepvisionlanguage} & $\checkmark$ & -- \\
SpanVLA \citep{zhou2026spanvlaefficientactionbridging} & -- & $\checkmark$ \\
UniDriveVLA \citep{li2026unidrivevlaunifyingunderstandingperception} & $\checkmark$ & -- \\
\midrule
\multicolumn{3}{@{}l}{\textbf{Manipulation}} \\
$\pi_{0.5}$ \citep{intelligence2025pi_05} & $\checkmark$ & $\checkmark$ \\
MolmoAct2 \citep{fang2026molmoact2actionreasoningmodels} & $\checkmark$ & $\checkmark$ \\
\bottomrule
\end{tabular}
\end{table}

Backbone supervision also affects whether pretrained capabilities survive VLA adaptation. End-to-end fine-tuning can distort useful features, motivating parameter-efficient adaptation, language-aligned action representations, and continued vision--language co-training \citep{kumar2022finetuningdistortpretrainedfeatures,hancock2025actionslanguagefinetuningvlms,grover2025enhancinggeneralizationvisionlanguageactionmodels,zha2026laplanguageactionpretrainingenables}. These approaches address how the VLM is trained, but leave unresolved a key interaction: whether autoregressive supervision over a discrete representation of the continuous action target affects the benefit of allowing gradients from the continuous expert to flow into the VLM backbone.

\textbf{Gradient routing and training recipes.}
$\pi_{0.5}$, MolmoAct2, and Alpamayo-R1 adopt Knowledge Insulation: they train the continuous action expert without propagating its gradients into the VLM backbone, while updating the VLM through autoregressive action-token and semantic objectives \citep{driess2026knowledge,intelligence2025pi_05,fang2026molmoact2actionreasoningmodels,wang2025alpamayo}. Within this framework, $\pi_{0.5}$ and MolmoAct2 use a two-stage schedule. The VLM is first adapted through autoregressive prediction of tokenized actions and semantic targets and the continuous expert is introduced afterward. Variants of this staged, insulated training recipe are common in recent VLAs; Table~\ref{tab:insulated_vla_recipes} summarizes representative works that insulate or freeze the VLM during action-expert training. By contrast, Wall-OSS-0.5 favors coupled optimization, allowing gradients from the continuous expert to update the VLM while retaining discrete-action and multimodal objectives \citep{yu2026walloss05technicalreport}. Although these works motivate different gradient-routing choices at a shared VLM--expert interface, their results are not directly comparable because they also differ in training recipes, initialization, and data.

The growing use of this hybrid autoregressive--continuous MoT framework has not resolved how models should be trained. $\pi_{0.7}$ scales a FAST-supervised version of the recipe; MolmoAct2 changes its gradient routing between broad post-training and embodiment-specific fine-tuning; and APT finds that action-expert preparation changes the relative performance of insulated and coupled training \citep{intelligence2026pi07steerablegeneralistrobotic,fang2026molmoact2actionreasoningmodels,xu2026aptactionexpertpretraining}. Together with the contrasting Wall-OSS-0.5 results, these findings suggest that the preferred gradient routing depends on the surrounding training recipe. Related multimodal-pretraining studies similarly show that modalities can reinforce or interfere with one another depending on parameter sharing and training order \citep{han2026physicsmultimodalpretrainingknowledge}. Prior work therefore establishes gradient insulation as a consequential design choice, but does not isolate its interaction with autoregressive action supervision. We vary these decisions independently while holding semantic supervision and the VLM--expert architecture fixed.

\section{Methodology}
\label{sec:method}

In this work, we test whether supervision on discrete future-action tokens changes the effect of propagating gradients from a continuous action expert into the VLM. We first study this question in autonomous driving, using the Alpamayo VLA architecture \citep{wang2025alpamayo} and the hybrid autoregressive and flow-matching recipe used by $\pi_{0.5}$ and Knowledge Insulation \citep{intelligence2025pi_05,driess2026knowledge}. We vary future-action-token supervision and expert-to-VLM gradient flow independently, while holding the architecture and semantic supervision fixed.

\begin{figure}[t]
    \centering
    \includegraphics[width=1.00\linewidth]{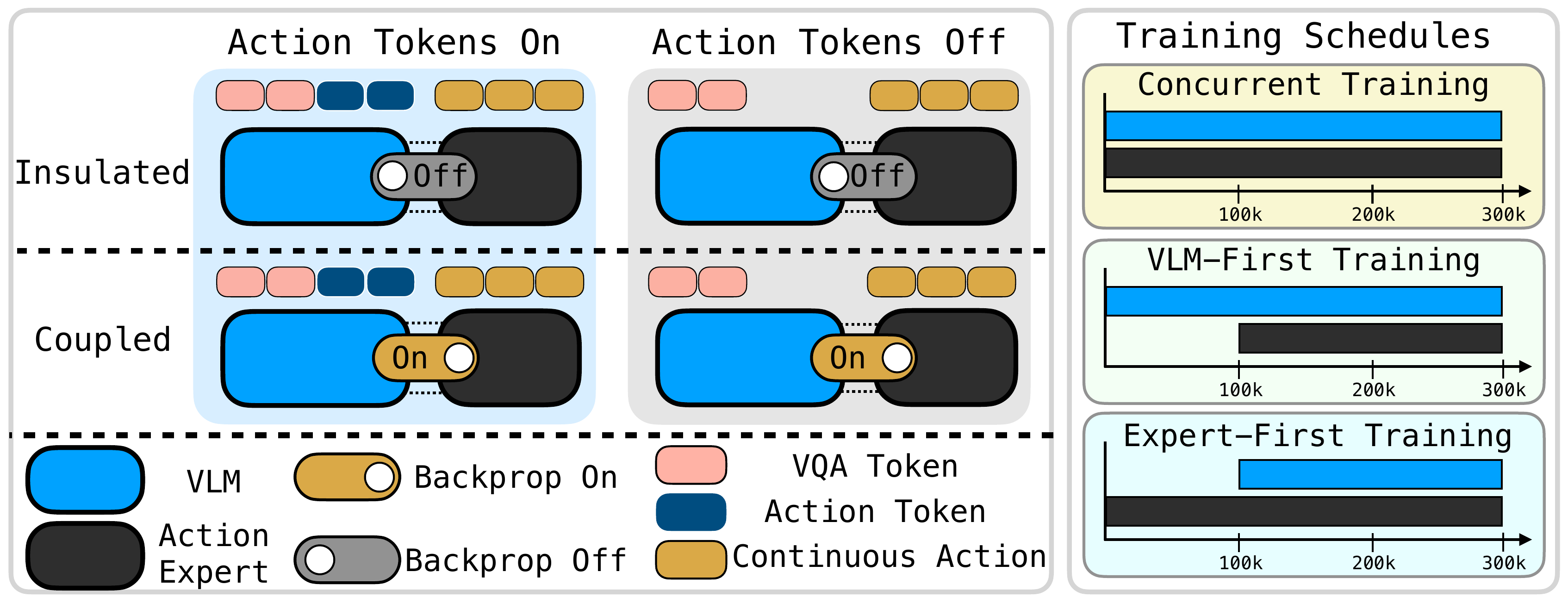}
    \caption{\textbf{Action-token supervision, gradient routing, and training schedules.} \textbf{Left}: we cross future-action-token supervision with coupled or insulated expert-to-VLM gradients. \textbf{Right}: bars show when the VLM (blue) and action expert (black) are trained. Concurrent training starts both modules together; VLM-first and expert-first training start one module after 100k steps of training the other module. All schedules run for 300k total steps in autonomous driving experiments.}
    \label{fig:methodlogy}

\end{figure}

\textbf{Mixture-of-Transformers VLA.}
The driving model pairs a Qwen3-VL-2B-Instruct \citep{yang2025qwen3technicalreport} backbone with a same-width, non-causal flow-matching action expert. At each transformer layer, the expert attends to key--value representations produced by the corresponding VLM layer, while the VLM does not attend to the expert's continuous action inputs. The expert predicts $K$ future control pairs, each comprising acceleration and curvature. Starting from the current ego state, these controls are integrated using a discretized unicycle model to recover position and heading waypoints for low-level control (see \ref{app:driving_architecture} for further details). We retain this architecture for every experiment.

Training includes trajectory examples with and without reasoning annotations. When present, the VLM predicts a concise, scene-grounded rationale identifying relevant road users or scene features and the intended driving response, followed by the future-trajectory start token. In action-token-supervised conditions, the VLM also predicts the subsequent future-trajectory (action) tokens. At evaluation, the VLM always generates a rationale and stops decoding at the start token. The action expert then predicts continuous controls conditioned on the resulting VLM K/V cache. Thus, future-action tokens are never decoded at evaluation and, when supervised, serve only as an auxiliary training signal shaping VLM representations.

The VLM is trained on the standard cross-entropy (CE) objective, while the expert is trained through rectified flow matching. The joint objective is
$\mathcal{L}=\mathcal{L}_{\mathrm{CE}}+\mathcal{L}_{\mathrm{flow}}$, where
\begin{align}
\mathcal{L}_{\mathrm{CE}}
&=
-\mathbb{E}_{(x,y)\sim\mathcal{D}}
\left[
\frac{1}{N}
\left(
\sum_{i\in\mathcal{I}_{\mathrm{base}}}
\log p_{\theta}(y_i\mid x,y_{<i})
+
\mathbf{1}_{\mathrm{action}}
\sum_{i\in\mathcal{I}_{\mathrm{action}}}
\log p_{\theta}(y_i\mid x,y_{<i})
\right)
\right],
\label{eq:ce_loss}
\\
\mathcal{L}_{\mathrm{flow}}
&=
\mathbb{E}_{(x,a)\sim\mathcal{D}_{\mathrm{action}},\,t,\,\epsilon}
\left[
\left\|
v_{\phi}(a_t,t;\widetilde{H}_{\theta})-(a-\epsilon)
\right\|_2^2
\right],
\qquad
a_t=t a+(1-t)\epsilon.
\label{eq:flow_loss}
\end{align} 
We sample $b\sim\operatorname{Beta}(1.5,1.0)$, set $t=0.999(1-b)$, and sample $\epsilon\sim\mathcal{N}(0,I)$. Parameters $\theta$ and $\phi$ belong to the VLM and action expert, respectively; $x$ is the current timestep's visual observations, language context, and past trajectory history, $y$ is the autoregressive target sequence, and $a$ is the continuous trajectory. The term $\widetilde{H}_{\theta}$ denotes the VLM K/V representations supplied to the expert. For coupling, $\widetilde{H}_{\theta}=H_{\theta}$; for insulation, $\widetilde{H}_{\theta}=\operatorname{sg}(H_{\theta})$, where $\operatorname{sg}$ passes the representations unchanged to the expert but blocks gradients from the flow loss to the VLM \citep{driess2026knowledge}. The set $\mathcal{I}_{\mathrm{base}}$ contains always-supervised token, including VQA answers, driving rationales, and future-trajectory delimiters, while $\mathcal{I}_{\mathrm{action}}$ contains future-trajectory value tokens. The indicator $\mathbf{1}_{\mathrm{action}}$ enables or disables direct CE supervision on those value tokens. Both loss terms have unit weight.

Crucially, the discrete action tokens and continuous expert encode the \emph{same future trajectory}. Disabling action-token supervision masks CE only on future-action values, which remain teacher-forced; trajectory delimiters remain supervised, and past-trajectory tokens remain in the input context. We fix $N$ at the pre-mask count of supervised token positions, preserving the weight of other CE terms, and exclude future-action token states from the expert's K/V context.

\textbf{Gradient routing.}
Gradient routing changes only the backward pass through the layerwise K/V interface. In the \emph{coupled} condition, the flow loss propagates through this interface and updates both the expert and the VLM. In the \emph{insulated} condition, these gradients are stopped before entering the VLM. Stop-gradient leaves the forward computation unchanged for fixed parameters, so both conditions use the same expert conditioning mechanism and flow-matching target. Figure~\ref{fig:methodlogy} summarizes which target signals update the VLM in each condition. VQA, reasoning, and structural-token supervision remain active in every condition; past-trajectory tokens remain in the input context.

\textbf{Training schedules.}
To test whether training order changes the interaction between action-token supervision and gradient routing, we evaluate all four conditions under three training schedules (see Figure ~\ref{fig:methodlogy}). In \emph{concurrent training}, the pretrained VLM and a randomly initialized action expert are trained together from the outset. In \emph{expert-first training}, motivated by APT's expert-preparation strategy \citep{xu2026aptactionexpertpretraining}, the expert is trained for 100k steps against a frozen VLM before joint training begins. Neither action-token supervision nor gradient routing affects this first stage because the VLM is frozen. In \emph{VLM-first training}, reflecting the staged adaptation used by $\pi_{0.5}$ and MolmoAct2 \citep{intelligence2025pi_05,fang2026molmoact2actionreasoningmodels}, the VLM is trained alone for 100k steps before a newly initialized expert is introduced. Action-token supervision varies during this first stage, while gradient routing applies only after the expert is introduced. Experts share the same initialization across conditions.

All joint-training stages use the same mixture of trajectory and vision--language batches, with trajectory-bearing batches constituting approximately $48.5\%$ of the data. The VLM receives an update on every joint-training step, whereas the expert receives one only on trajectory-bearing steps. During the VLM-first warm-up, tokens-off models use only vision--language data, whereas tokens-on models also receive trajectory data. During the expert-first warm-up, every step updates the expert because it uses trajectory data exclusively. Thus, matching total optimizer steps across histories does not equalize component-level update budgets; additional details appear in Appendix~\ref{app:training_details}.

We compare checkpoints after 300k total optimizer steps, when validation performance had stabilized. Each schedule has the same total step budget: 300k concurrent steps or a 100k single-component stage followed by 200k joint steps. Within each schedule, the four conditions share the starting pretrained checkpoint, expert initialization, and optimizer schedule. Batch sampling is matched between routing conditions within each action-token setting.

Our main experiments evaluate these interventions in autonomous driving. In Section~\ref{sec:manipulation}, we repeat the concurrent-training experiment in closed-loop robot manipulation to test whether the findings extend beyond driving. Architecture, action-space, optimization, data, and flow-matching details appear in Appendix~\ref{app:method_details}.

\section{Experiments}
\label{sec:experiments}
In this section, we ask three questions: (\textbf{Q1}) Does action-token supervision change the most performative gradient routing mechanism? (\textbf{Q2}) Does pretraining the VLM or expert first change that answer? (\textbf{Q3}) Are findings consistent across application domains? We address these questions in the three results subsections below. 

\textbf{Open-loop driving.} We evaluate on the publicly released Physical AI Autonomous Vehicles dataset, specifically its nominal (named ``Public Golden") and out-of-distribution (``OOD") splits, which contain 1,434 and 273 scenes, respectively \citep{nvidia_physicalai_av_2026}. For each example, the policy samples six candidate trajectories. Our primary metric is average displacement error (ADE), the mean Euclidean distance between predicted and ground-truth positions over the future trajectory. We report ADE for the first sampled candidate. Minimum ADE (minADE) is the lowest ADE among the six candidates and serves as an oracle selection metric.

\textbf{Closed-loop driving.} We evaluate in AlpaSim, a closed-loop simulator with realistic sensor observations, vehicle dynamics, and traffic scenarios \citep{alpasim_2025}. Every condition is evaluated on the same 827 scenes with five stochastic rollouts per scene. We average rollouts within each scene to reduce evaluation variance.

\textbf{Driving rationales.} We use a frozen Lingo-Judge \citep{marcu2024lingoqavisualquestionanswering} classifier to compare generated reasoning-traces with reference rationales on the Public Golden and OOD benchmarks. The first-sample pass rate is the fraction of examples whose first generated rationale receives a score above 0.5. We also report the fraction for which any of six candidates passes and the mean score across candidates.

\textbf{Statistical analysis.} For open-loop ADE and first-sample rationale pass rates, we compute 95\% confidence intervals from 10,000 paired clip-cluster bootstrap resamples. Each resample draws whole clips with replacement and applies the same sampled clips to every condition. These intervals quantify sensitivity to the evaluated clips for fixed checkpoints.

\subsection{Concurrent Training}
\label{sec:concurrent_driving}

\begin{figure}[t]
    \centering
    \includegraphics[width=\linewidth]{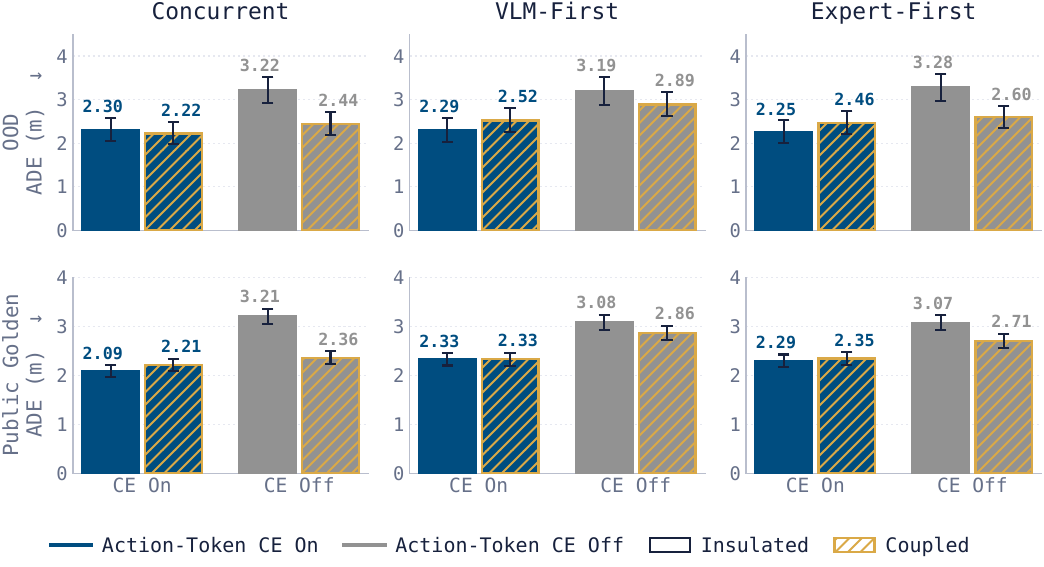}
    \caption{\textbf{Open-loop driving accuracy across training histories.} First-sample mean ADE on the Public Golden (1,434 examples) and OOD (273 examples) evaluation splits. Bars show the mean across scenes; error bars show 95\% clip-cluster bootstrap confidence intervals. Lower is better.}
    \label{fig:concurrent_driving_endpoints}
\end{figure}

\begin{table*}[t]
\centering
\scriptsize
\setlength{\tabcolsep}{3pt}
\renewcommand{\arraystretch}{1.02}
\caption{\textbf{Generated reasoning-trace quality.} First-sample Lingo-Judge pass rates for each model. In each example, the frozen Lingo-Judge classifier compares the reasoning-trace with a reference rationale \citep{marcu2024lingoqavisualquestionanswering}; an example passes when the judge score exceeds $0.5$. Brackets give 95\% clip-cluster bootstrap confidence intervals. Top-$k$ is the fraction for which any of six sampled rationales passes, and Avg-$k$ is the mean judge score across the six samples. Results are evaluated on Public Golden and OOD. Within each training schedule, the best metric is bolded. Higher is better.}

\label{tab:lingo_judge_300k}
\begin{tabular*}{\textwidth}{@{\extracolsep{\fill}}llccccccc@{}}
\toprule
& & &
\multicolumn{3}{c}{\textbf{Public Golden} ($N=1{,}434$)} &
\multicolumn{3}{c}{\textbf{OOD} ($N=273$)} \\
\cmidrule(lr){4-6}\cmidrule(l){7-9}
\textbf{Schedule} & \textbf{Routing} & \textbf{Action CE} &
\textbf{Top-1 [95\% CI]} & \textbf{Top-$k$} & \textbf{Avg-$k$} &
\textbf{Top-1 [95\% CI]} & \textbf{Top-$k$} & \textbf{Avg-$k$} \\
\midrule
Concurrent   & Insulated & On  & 0.647 [0.621,0.673] & 0.808 & \textbf{0.630} & 0.414 [0.355,0.474] & 0.659 & 0.466 \\
Concurrent   & Coupled   & On  & \textbf{0.649} [0.623,0.674] & \textbf{0.823} & 0.629 & \textbf{0.425} [0.364,0.487] & \textbf{0.689} & \textbf{0.471} \\
Concurrent   & Insulated & Off & 0.566 [0.539,0.593] & 0.767 & 0.573 & 0.363 [0.305,0.423] & 0.652 & 0.435 \\
Concurrent   & Coupled   & Off & 0.635 [0.610,0.661] & 0.811 & 0.622 & 0.348 [0.292,0.406] & 0.685 & 0.450 \\
\addlinespace[2pt]
VLM-First    & Insulated & On  & \textbf{0.639} [0.614,0.665] & 0.807 & \textbf{0.626} & 0.392 [0.335,0.452] & 0.652 & 0.462 \\
VLM-First    & Coupled   & On  & 0.635 [0.609,0.661] & \textbf{0.814} & 0.625 & 0.403 [0.344,0.464] & 0.641 & 0.458 \\
VLM-First    & Insulated & Off & 0.630 [0.603,0.657] & 0.800 & 0.616 & \textbf{0.447} [0.386,0.508] & \textbf{0.711} & \textbf{0.465} \\
VLM-First    & Coupled   & Off & 0.613 [0.585,0.639] & 0.798 & 0.611 & 0.396 [0.339,0.452] & 0.667 & 0.451 \\
\addlinespace[2pt]
Expert-First & Insulated & On  & 0.644 [0.618,0.670] & \textbf{0.819} & 0.626 & \textbf{0.458} [0.399,0.518] & \textbf{0.700} & \textbf{0.470} \\
Expert-First & Coupled   & On  & \textbf{0.649} [0.623,0.675] & 0.811 & \textbf{0.628} & 0.388 [0.331,0.447] & 0.659 & 0.453 \\
Expert-First & Insulated & Off & 0.620 [0.594,0.646] & 0.804 & 0.610 & 0.407 [0.348,0.466] & 0.659 & 0.450 \\
Expert-First & Coupled   & Off & 0.612 [0.585,0.639] & 0.798 & 0.611 & 0.392 [0.333,0.451] & 0.692 & 0.459 \\
\bottomrule
\end{tabular*}

\end{table*}

We first address Q1 in the simplest setting, where the VLM backbone and a randomly initialized action expert are trained together from the outset. Recall that all four conditions share the same data mixture, initialization, and optimization schedule.

\textbf{Open-loop driving.} Figure~\ref{fig:concurrent_driving_endpoints} shows that the effect of gradient routing depends on action-token supervision. With action-token supervision, insulation slightly lowers ADE on the Public Golden split, while the OOD routing difference is comparatively small. Without action-token supervision, the insulated model has substantially higher ADE in both evaluations. Relative to insulation, coupling reduces ADE by approximately 26\% on the Public Golden dataset and 24\% OOD.

This roughly one-quarter reduction in mean trajectory error is consistent with action-token supervision and expert gradients providing partially substitutable signals to the VLM. Action-token cross-entropy loss directly supervises the future trajectory even when expert gradients are blocked; without it, the flow loss is the only direct future-action objective that updates the VLM.

\textbf{Generated reasoning-trace quality.} We use the first-sample Lingo-Judge pass rate as a diagnostic of the language generated during concurrent training (Table~\ref{tab:lingo_judge_300k}). With action-token supervision, insulated and coupled models obtain nearly identical pass rates on Public Golden (64.7\% and 64.9\%, respectively). Without action-token supervision, the coupled model has a higher observed pass rate on Public Golden (63.5\% versus 56.6\%), while the OOD routing difference is small and points in the opposite direction. Insulation shows no consistent advantage in generated driving-rationale quality under the evaluated vision--language co-training recipe. 

\textbf{Closed-loop driving.} In concurrent training, coupling raises AlpaSim route progress from \(0.600\) to \(0.703\) with action-token supervision and from \(0.425\) to \(0.601\) without it (Table~\ref{tab:alpasim_progress_histories}). The latter is a 17.6 percentage-point increase, or about 41\% relative, on the same evaluation scenes. Coupling without tokens also reaches approximately the same route progress as insulated training with tokens. Thus, allowing the expert's loss to update the VLM yields a substantial gain in interactive driving, with the largest observed gain when action-token supervision is absent.

Thus, Q1 has a conditional answer: without action-token supervision, coupling substantially improves open- and closed-loop driving; with supervision, it still improves closed-loop progress, while the open-loop difference is small. The most important factor is that the VLM receives a direct action learning signal, supplied by either action-token supervision or expert-gradient coupling.

\subsection{Dependence on Training History}
\label{sec:staged_driving}

To address Q2, we next test whether the supervision--routing pattern depends on training history by first adapting either the action expert or the VLM for 100k steps. We align staged and concurrent checkpoints by total optimizer steps; component-level gradient exposure nevertheless differs across histories by design (see Table~\ref{tab:update_budget} in the Appendix).

\textbf{Open-loop driving.} The supervision--routing pattern observed under concurrent training persists when either module is adapted first (Figure~\ref{fig:concurrent_driving_endpoints}). Across both evaluation datasets, insulation increases mean ADE without action-token supervision in every history, while its effect with supervision is smaller and depends on training order. Thus, training order changes the size of the routing effect but does not remove its dependence on action-token supervision.

\textbf{Generated reasoning-trace quality.} The staged Lingo-Judge results do not mirror the trajectory-ADE pattern (Table~\ref{tab:lingo_judge_300k}). Without action-token supervision, insulated models have higher observed first-sample pass rates on Public Golden under both staged schedules, even though coupled models have lower ADE. Routing differences on OOD vary. Thus, improvements in continuous trajectory prediction do not necessarily coincide with improvements in generated reasoning traces; neither gradient routing choice uniformly favors both outcomes.

\textbf{Closed-loop driving.} We evaluate both staged histories on the same AlpaSim scenes using the concurrent-training protocol from Section~\ref{sec:concurrent_driving}. Table~\ref{tab:alpasim_progress_histories} shows that coupling increases observed route progress without action-token supervision after both Expert-First and VLM-First training. With supervision, coupling raises progress after Expert-First training but makes little difference after VLM-First training. Coupling without tokens roughly matches insulated training with tokens in the Expert-First history, but falls below it in the VLM-First history. Concurrent coupled training with tokens achieves the highest observed progress overall (\(0.703\)). Full staged competence and safety results appear in Appendix Tables~\ref{tab:alpasim_expert_first_300k} and~\ref{tab:alpasim_vlm_first_300k}.

\begin{table}[!htbp]
    \centering

    \small
    \captionsetup{width=\linewidth}
    \setlength{\tabcolsep}{6pt}
    \renewcommand{\arraystretch}{1.18}
    \caption{\textbf{Closed-loop route progress across training histories.} Mean AlpaSim route progress on the same 827 scenes, with five stochastic rollouts per scene averaged within each scene. The best value in each row is bolded and the second-best is underlined. Higher is better.}
    \label{tab:alpasim_progress_histories}
    \begin{tabular}{@{}lrrrr@{}}
        \toprule
        & \multicolumn{2}{c}{\textbf{Insulated}} & \multicolumn{2}{c}{\textbf{Coupled}} \\
        \cmidrule(lr){2-3}\cmidrule(l){4-5}
        \textbf{History}
            & \shortstack{\textbf{Tokens}\\\textbf{on}}
            & \shortstack{\textbf{Tokens}\\\textbf{off}}
            & \shortstack{\textbf{Tokens}\\\textbf{on}}
            & \shortstack{\textbf{Tokens}\\\textbf{off}} \\
        \midrule
        Concurrent & 0.600 & 0.425 & \textbf{0.703} & \underline{0.601} \\
        VLM-First & \textbf{0.648} & 0.555 & \underline{0.646} & 0.587 \\
        Expert-First & 0.597 & 0.538 & \textbf{0.625} & \underline{0.604} \\
        \bottomrule
    \end{tabular}

\end{table}

Thus, for Q2, training order changes the size of the routing effect, but coupling improves open-loop accuracy without action-token supervision under every schedule. Importantly, \emph{the greatest absolute performance in action-prediction across all training histories is achieved by concurrent training}, which stands in contrast to nominal VLM-First recipes adopted by prior works \citep{intelligence2025pi_05, fang2026molmoact2actionreasoningmodels}. 

\subsection{Transfer to Robot Manipulation}
\label{sec:manipulation}

Finally, to address Q3, we repeat the concurrent training recipe from Section ~\ref{sec:method} in robot manipulation with two model families: a $\pi_{0.5}$ re-implementation \cite{intelligence2025pi_05} and MolmoAct2 \cite{fang2026molmoact2actionreasoningmodels}. The four training conditions vary FAST supervision and expert-to-VLM gradient routing while sharing the data mixture, initialization procedure, and optimization schedule to isolate the effect of action-token prediction and gradient insulation. 

Our $\pi_{0.5}$ re-implementation pairs a PaliGemma-3B VLM with a flow-matching action expert \citep{beyer2024paligemmaversatile3bvlm}. We use OpenPi \citep{black2024pi_0} with the FAST implementation from \citet{zha2026laplanguageactionpretrainingenables}, and train for 400k steps. Batches are divided equally between VQA and robot data. The VQA mixture comprises VQAv2 question answering (35\%) \citep{goyal2017makingvvqamatter}, COCO captioning (20\%) \citep{chen2015microsoftcococaptionsdata} and localization (20\%) \citep{lin2015microsoftcococommonobjects}, and Visual Genome question answering (15\%) and region tasks (10\%) \citep{krishna2016visualgenomeconnectinglanguage}. Robot data comprises 90\% DROID \citep{khazatsky2025droidlargescaleinthewildrobot} and 10\% PolaRiS training data \citep{jain2025polaris}.

For MolmoAct2 \citep{fang2026molmoact2actionreasoningmodels}, we initialize all four conditions from the official Molmo2-ER checkpoint, with fresh action experts and train concurrently for 300k steps. We reconstruct its published mixture from available public sources, retaining the 90\% robot / 10\% multimodal split. Each checkpoint then receives 1k steps of adaptation on 90\% DROID and 10\% PolaRiS training data. Architecture, attention interfaces, data mixtures, optimization, and adaptation settings are detailed in Appendix~\ref{app:manipulation_setup}.

\begin{table}[!htbp]
    \centering
    \footnotesize

    \captionsetup{width=\linewidth}
    \caption{\textbf{Closed-loop manipulation progress.}
    Equal-weight mean normalized \emph{task progress} across three representative PolaRiS tasks. Bold indicates the best result among our four conditions within each model family. The final row provides additional references: the published $\pi_{0.5}$ result is from CoVeR \citep{kwok2026scalingverificationeffectivescaling} and uses its evaluation protocol; the MolmoAct2 result uses its released checkpoint with our own fine-tuning and evaluation scheme. Higher is better.}
    \label{tab:polaris}
    \setlength{\tabcolsep}{3pt}
    \renewcommand{\arraystretch}{1.1}
    \begin{tabular}{@{}lrr@{}}
        \toprule
        \textbf{Condition} & $\boldsymbol{\pi_{0.5}}$ & \textbf{MolmoAct2} \\
        \midrule
        Insulated / FAST on  & 19.5\% & 22.0\% \\
        Coupled / FAST on    & 30.7\% & 24.3\% \\
        Insulated / FAST off & 10.3\% & 14.5\% \\
        Coupled / FAST off   & \textbf{32.9\%} & \textbf{36.5\%} \\
        \midrule
        Insulated / FAST on & 40.0\% & 32.9\% \\
        \bottomrule
    \end{tabular}

\end{table}

We evaluate both model families in the same three PolaRiS environments, which support comparisons from matched initial states \citep{jain2025polaris}. \texttt{BlockStackKitchen} requires placing and stacking blocks on a green tray; \texttt{FoodBussing} requires putting food items into a bowl; and \texttt{PanClean} requires using a sponge on a frying pan in a cluttered kitchen. We run 50 rollouts per condition and environment from matched initial states. Because full task completion is sparse, our primary metric is normalized task progress, averaged equally across the three environments. Normalized task progress is the fraction of rubric criteria achieved during an episode, with credit retained once a criterion is satisfied (Appendix~\ref{app:additional_manipulation_results}).

For our $\pi_{0.5}$ re-implementation, the observed gain from coupling is larger without FAST: progress rises from 10.3\% to 32.9\%, compared with 19.5\% to 30.7\% with FAST (Table~\ref{tab:polaris}). Coupled training without FAST achieves the highest mean progress among our four conditions, about 1.7 times that of insulated training with FAST. For context, CoVeR reports 40.0\% progress for a published $\pi_{0.5}$ checkpoint under its PolaRiS evaluation protocol \citep{kwok2026scalingverificationeffectivescaling}. That checkpoint was pretrained on more than 10,000 hours of robot data \citep{intelligence2025pi_05}, whereas our robot training draws primarily from DROID, a 350-hour dataset \citep{khazatsky2025droidlargescaleinthewildrobot}.

MolmoAct2 exhibits the same qualitative pattern: coupling increases progress from 22.0\% to 24.3\% with FAST and from 14.5\% to 36.5\% without FAST. As before, coupled training without action token prediction achieves the highest mean progress among our four training strategies. For context, the released MolmoAct2 checkpoint reaches 32.9\% after our adaptation and evaluation protocol (Table~\ref{tab:polaris}). Unlike our concurrent training recipe, the released checkpoint follows a VLM-First procedure that trains discrete action prediction with FAST before introducing the continuous expert \citep{fang2026molmoact2actionreasoningmodels}. Differences in training history and the VLM training mixture may contribute to the performance gap between our concurrent insulated model with FAST and the released checkpoint. The architecture, attention interfaces, training, and adaptation details appear in Appendix~\ref{app:manipulation_setup}. Thus, the answer to Q3 is yes: coupling improves observed closed-loop manipulation progress with and without FAST supervision in both model families, with larger gains when FAST supervision is absent.

\section{Discussion and Conclusion}
\label{sec:discussion}
Insulating the VLM from action-expert gradients is often motivated by preserving its pretrained capabilities. Our findings show that the consequences of this choice depend on the other signals available to the VLM. Action-token supervision directly trains it to predict future actions, which helps explain why blocking expert gradients has little effect on open-loop driving error when that supervision is present. Without action tokens, insulation substantially worsens trajectory prediction across evaluation sets and training schedules. Yet under concurrent training, coupling improves closed-loop route progress even with action-token supervision, and the same pattern appears in manipulation. Routing effects also vary across safety metrics: with action-token supervision, coupled staged training has higher observed at-fault collision rates, despite higher or similar route progress (Appendix Tables~\ref{tab:alpasim_expert_first_300k} and~\ref{tab:alpasim_vlm_first_300k}). 

These results are consistent with action-token supervision and expert gradients providing partially substitutable action-learning signals. Insulation shows no consistent advantage in generated driving-rationale quality under the evaluated vision--language co-training recipe, although this finding does not establish broader VLM capability retention. Moreover, concurrently training the VLM and action expert from the outset is a strong, simple starting point. Our results suggest that gradient routing should be chosen jointly with action-token supervision and the training schedule, with its benefits assessed for the intended evaluation setting. In summary, \emph{the value of VLM insulation depends on action supervision, training schedule, and evaluation setting, thereby calling into question its use as a default VLA training strategy.}

\textbf{Limitations.} We study only one driving model family and two manipulation model families, making our study somewhat limited in architectural diversity. Moreover, our closed-loop evaluations use only simulated driving scenes and simulated manipulation tasks. We do not test the manipulation policies on physical robots, which is an important next step for assessing whether the observed gains transfer to real-world control. Additionally, while our bootstrap intervals capture variation across evaluation clips for fixed checkpoints, compute constraints limited us to one training run per condition, leaving training-seed variation unmeasured. Finally, our study is empirical: we characterize the interaction between action-token supervision and gradient routing, but do not establish the mechanisms underlying the observed performance differences.

\textbf{Future Work.} Understanding how action-token supervision and expert gradients shape the VLM's representations and learning dynamics is an important next step toward explaining when insulation helps or hurts. We are also excited to test whether this interaction extends to more embodiments, world-action models, and action-conditioned video world models.

\bibliographystyle{paper}
\bibliography{paper}

@article{black2024pi_0,
  title={{$\pi_0$}: A Vision-Language-Action Flow Model for General Robot Control},
  author={Black, Kevin and Brown, Noah and Driess, Danny and Esmail, Adnan and Equi, Michael and Finn, Chelsea and Fusai, Niccolo and Groom, Lachy and Hausman, Karol and Ichter, Brian and others},
  journal={arXiv preprint arXiv:2410.24164},
  year={2024}
}

@article{liang2024mixture,
  title={Mixture-of-transformers: A sparse and scalable architecture for multi-modal foundation models},
  author={Liang, Weixin and Yu, Lili and Luo, Liang and Iyer, Srinivasan and Dong, Ning and Zhou, Chunting and Ghosh, Gargi and Lewis, Mike and Yih, Wen-tau and Zettlemoyer, Luke and others},
  journal={arXiv preprint arXiv:2411.04996},
  year={2024}
}

@article{intelligence2025pi_05,
  title={{$\pi_{0.5}$}: A Vision-Language-Action Model with Open-World Generalization},
  author={{Physical Intelligence} and Black, Kevin and Brown, Noah and Darpinian, James and Dhabalia, Karan and Driess, Danny and Esmail, Adnan and Equi, Michael and Finn, Chelsea and Fusai, Niccolo and others},
  journal={arXiv preprint arXiv:2504.16054},
  year={2025}
}

@article{driess2026knowledge,
  title={Knowledge insulating vision-language-action models: Train fast, run fast, generalize better},
  author={Driess, Danny and Springenberg, Jost and Ichter, Brian and Yu, Lili and Li-Bell, Adrian and Pertsch, Karl and Ren, Allen and Walke, Homer and Vuong, Quan and Shi, Lucy Xiaoyang and others},
  journal={Advances in Neural Information Processing Systems},
  volume={38},
  pages={102867--102888},
  year={2026}
}

@article{wang2025alpamayo,
  title={{Alpamayo-R1}: Bridging reasoning and action prediction for generalizable autonomous driving in the long tail},
  author={NVIDIA and Wang, Yan and Luo, Wenjie and Bai, Junjie and Cao, Yulong and Che, Tong and Chen, Ke and Chen, Yuxiao and Diamond, Jenna and Ding, Yifan and Ding, Wenhao and others},
  journal={arXiv preprint arXiv:2511.00088},
  year={2025}
}

@article{pertsch2025fast,
  title={Fast: Efficient action tokenization for vision-language-action models},
  author={Pertsch, Karl and Stachowicz, Kyle and Ichter, Brian and Driess, Danny and Nair, Suraj and Vuong, Quan and Mees, Oier and Finn, Chelsea and Levine, Sergey},
  journal={arXiv preprint arXiv:2501.09747},
  year={2025}
}

@misc{yu2026walloss05technicalreport,
      title={Wall-OSS-0.5 Technical Report}, 
      author={Ryan Yu and Pushi Zhang and Starrick Liu and Brae Liu and Miracle Kang and Shalfun Li and Lights Shi and Ellie Ma and Ping Yang and Chris Pan and Jerry Chen and Dongxiu Liu and Rain Sun and Miles Guo and Byron Zhang and Hugo Zhou and Zach Xu and Vincent Chen and Harrison Huang and James Wang and Dance Kuzi and Andy Zhai and Hang Su and Roy Gan and Lucy Liang and Hao Wang and Qian Wang},
      year={2026},
      eprint={2605.30877},
      archivePrefix={arXiv},
      primaryClass={cs.RO},
      url={https://arxiv.org/abs/2605.30877}, 
}

@misc{brohan2023rt2visionlanguageactionmodelstransfer,
      title={RT-2: Vision-Language-Action Models Transfer Web Knowledge to Robotic Control}, 
      author={Anthony Brohan and Noah Brown and Justice Carbajal and Yevgen Chebotar and Xi Chen and Krzysztof Choromanski and Tianli Ding and Danny Driess and Avinava Dubey and Chelsea Finn and Pete Florence and Chuyuan Fu and Montse Gonzalez Arenas and Keerthana Gopalakrishnan and Kehang Han and Karol Hausman and Alexander Herzog and Jasmine Hsu and Brian Ichter and Alex Irpan and Nikhil Joshi and Ryan Julian and Dmitry Kalashnikov and Yuheng Kuang and Isabel Leal and Lisa Lee and Tsang-Wei Edward Lee and Sergey Levine and Yao Lu and Henryk Michalewski and Igor Mordatch and Karl Pertsch and Kanishka Rao and Krista Reymann and Michael Ryoo and Grecia Salazar and Pannag Sanketi and Pierre Sermanet and Jaspiar Singh and Anikait Singh and Radu Soricut and Huong Tran and Vincent Vanhoucke and Quan Vuong and Ayzaan Wahid and Stefan Welker and Paul Wohlhart and Jialin Wu and Fei Xia and Ted Xiao and Peng Xu and Sichun Xu and Tianhe Yu and Brianna Zitkovich},
      year={2023},
      eprint={2307.15818},
      archivePrefix={arXiv},
      primaryClass={cs.RO},
      url={https://arxiv.org/abs/2307.15818}, 
}

@misc{kim2024openvlaopensourcevisionlanguageactionmodel,
      title={OpenVLA: An Open-Source Vision-Language-Action Model}, 
      author={Moo Jin Kim and Karl Pertsch and Siddharth Karamcheti and Ted Xiao and Ashwin Balakrishna and Suraj Nair and Rafael Rafailov and Ethan Foster and Grace Lam and Pannag Sanketi and Quan Vuong and Thomas Kollar and Benjamin Burchfiel and Russ Tedrake and Dorsa Sadigh and Sergey Levine and Percy Liang and Chelsea Finn},
      year={2024},
      eprint={2406.09246},
      archivePrefix={arXiv},
      primaryClass={cs.RO},
      url={https://arxiv.org/abs/2406.09246}, 
}

@misc{li2024cogactfoundationalvisionlanguageactionmodel,
      title={CogACT: A Foundational Vision-Language-Action Model for Synergizing Cognition and Action in Robotic Manipulation}, 
      author={Qixiu Li and Yaobo Liang and Zeyu Wang and Lin Luo and Xi Chen and Mozheng Liao and Fangyun Wei and Yu Deng and Sicheng Xu and Yizhong Zhang and Xiaofan Wang and Bei Liu and Jianlong Fu and Jianmin Bao and Dong Chen and Yuanchun Shi and Jiaolong Yang and Baining Guo},
      year={2024},
      eprint={2411.19650},
      archivePrefix={arXiv},
      primaryClass={cs.RO},
      url={https://arxiv.org/abs/2411.19650}, 
}

@misc{nvidia2025gr00tn1openfoundation,
      title={GR00T N1: An Open Foundation Model for Generalist Humanoid Robots}, 
      author={NVIDIA and : and Johan Bjorck and Fernando Castañeda and Nikita Cherniadev and Xingye Da and Runyu Ding and Linxi "Jim" Fan and Yu Fang and Dieter Fox and Fengyuan Hu and Spencer Huang and Joel Jang and Zhenyu Jiang and Jan Kautz and Kaushil Kundalia and Lawrence Lao and Zhiqi Li and Zongyu Lin and Kevin Lin and Guilin Liu and Edith Llontop and Loic Magne and Ajay Mandlekar and Avnish Narayan and Soroush Nasiriany and Scott Reed and You Liang Tan and Guanzhi Wang and Zu Wang and Jing Wang and Qi Wang and Jiannan Xiang and Yuqi Xie and Yinzhen Xu and Zhenjia Xu and Seonghyeon Ye and Zhiding Yu and Ao Zhang and Hao Zhang and Yizhou Zhao and Ruijie Zheng and Yuke Zhu},
      year={2025},
      eprint={2503.14734},
      archivePrefix={arXiv},
      primaryClass={cs.RO},
      url={https://arxiv.org/abs/2503.14734}, 
}

@misc{zhou2024transfusionpredicttokendiffuse,
      title={Transfusion: Predict the Next Token and Diffuse Images with One Multi-Modal Model}, 
      author={Chunting Zhou and Lili Yu and Arun Babu and Kushal Tirumala and Michihiro Yasunaga and Leonid Shamis and Jacob Kahn and Xuezhe Ma and Luke Zettlemoyer and Omer Levy},
      year={2024},
      eprint={2408.11039},
      archivePrefix={arXiv},
      primaryClass={cs.AI},
      url={https://arxiv.org/abs/2408.11039}, 
}

@misc{chi2024diffusionpolicyvisuomotorpolicy,
      title={Diffusion Policy: Visuomotor Policy Learning via Action Diffusion}, 
      author={Cheng Chi and Zhenjia Xu and Siyuan Feng and Eric Cousineau and Yilun Du and Benjamin Burchfiel and Russ Tedrake and Shuran Song},
      year={2024},
      eprint={2303.04137},
      archivePrefix={arXiv},
      primaryClass={cs.RO},
      url={https://arxiv.org/abs/2303.04137}, 
}

@misc{kim2025finetuningvisionlanguageactionmodelsoptimizing,
      title={Fine-Tuning Vision-Language-Action Models: Optimizing Speed and Success}, 
      author={Moo Jin Kim and Chelsea Finn and Percy Liang},
      year={2025},
      eprint={2502.19645},
      archivePrefix={arXiv},
      primaryClass={cs.RO},
      url={https://arxiv.org/abs/2502.19645}, 
}

@misc{liu2025hybridvlacollaborativediffusionautoregression,
      title={HybridVLA: Collaborative Diffusion and Autoregression in a Unified Vision-Language-Action Model}, 
      author={Jiaming Liu and Hao Chen and Pengju An and Zhuoyang Liu and Renrui Zhang and Chenyang Gu and Xiaoqi Li and Ziyu Guo and Sixiang Chen and Mengzhen Liu and Chengkai Hou and Mengdi Zhao and KC alex Zhou and Pheng-Ann Heng and Shanghang Zhang},
      year={2025},
      eprint={2503.10631},
      archivePrefix={arXiv},
      primaryClass={cs.CV},
      url={https://arxiv.org/abs/2503.10631}, 
}

@misc{hancock2025actionslanguagefinetuningvlms,
      title={Actions as Language: Fine-Tuning VLMs into VLAs Without Catastrophic Forgetting}, 
      author={Asher J. Hancock and Xindi Wu and Lihan Zha and Olga Russakovsky and Anirudha Majumdar},
      year={2025},
      eprint={2509.22195},
      archivePrefix={arXiv},
      primaryClass={cs.RO},
      url={https://arxiv.org/abs/2509.22195}, 
}

@misc{zha2026laplanguageactionpretrainingenables,
      title={LAP: Language-Action Pre-Training Enables Zero-shot Cross-Embodiment Transfer}, 
      author={Lihan Zha and Asher J. Hancock and Mingtong Zhang and Tenny Yin and Yixuan Huang and Dhruv Shah and Allen Z. Ren and Anirudha Majumdar},
      year={2026},
      eprint={2602.10556},
      archivePrefix={arXiv},
      primaryClass={cs.RO},
      url={https://arxiv.org/abs/2602.10556}, 
}

@misc{grover2025enhancinggeneralizationvisionlanguageactionmodels,
      title={Enhancing Generalization in Vision-Language-Action Models by Preserving Pretrained Representations}, 
      author={Shresth Grover and Akshay Gopalkrishnan and Bo Ai and Henrik I. Christensen and Hao Su and Xuanlin Li},
      year={2025},
      eprint={2509.11417},
      archivePrefix={arXiv},
      primaryClass={cs.RO},
      url={https://arxiv.org/abs/2509.11417}, 
}

@misc{liu2026orderedactiontokensvisuomotor,
      title={Ordered Action Tokens for Visuomotor Policy Learning}, 
      author={Chaoqi Liu and Yue Zhao and Haonan Chen and Xiaoshen Han and Jiawei Gao and Ehsan Adeli and Yilun Du},
      year={2026},
      eprint={2607.21670},
      archivePrefix={arXiv},
      primaryClass={cs.RO},
      url={https://arxiv.org/abs/2607.21670}, 
}

@misc{xu2026aptactionexpertpretraining,
      title={APT: Action Expert Pretraining Improves Instruction Generalization of Vision-Language-Action Policies}, 
      author={Kechun Xu and Zhenjie Zhu and Anzhe Chen and Rong Xiong and Yue Wang},
      year={2026},
      eprint={2606.12366},
      archivePrefix={arXiv},
      primaryClass={cs.RO},
      url={https://arxiv.org/abs/2606.12366}, 
}

@misc{intelligence2026pi07steerablegeneralistrobotic,
      title={${\pi}_{0.7}$: a Steerable Generalist Robotic Foundation Model with Emergent Capabilities}, 
      author={Physical Intelligence and Bo Ai and Ali Amin and Raichelle Aniceto and Ashwin Balakrishna and Greg Balke and Kevin Black and George Bokinsky and Shihao Cao and Thomas Charbonnier and Vedant Choudhary and Foster Collins and Ken Conley and Grace Connors and James Darpinian and Karan Dhabalia and Maitrayee Dhaka and Jared DiCarlo and Danny Driess and Michael Equi and Adnan Esmail and Yunhao Fang and Chelsea Finn and Catherine Glossop and Thomas Godden and Ivan Goryachev and Lachlan Groom and Haroun Habeeb and Hunter Hancock and Karol Hausman and Gashon Hussein and Victor Hwang and Brian Ichter and Connor Jacobsen and Szymon Jakubczak and Rowan Jen and Tim Jones and Gregg Kammerer and Ben Katz and Liyiming Ke and Mairbek Khadikov and Chandra Kuchi and Marinda Lamb and Devin LeBlanc and Brendon LeCount and Sergey Levine and Xinyu Li and Adrian Li-Bell and Vladislav Lialin and Zhonglin Liang and Wallace Lim and Yao Lu and Enyu Luo and Vishnu Mano and Nandan Marwaha and Aikys Mongush and Liam Murphy and Suraj Nair and Tyler Patterson and Karl Pertsch and Allen Z. Ren and Gavin Schelske and Charvi Sharma and Baifeng Shi and Lucy Xiaoyang Shi and Laura Smith and Jost Tobias Springenberg and Kyle Stachowicz and Will Stoeckle and Jiaming Tang and Jimmy Tanner and Shalom Tekeste and Marcel Torne and Kyle Vedder and Quan Vuong and Anna Walling and Haohuan Wang and Jason Wang and XuDong Wang and Chris Whalen and Samuel Whitmore and Blake Williams and Charles Xu and Sukwon Yoo and Lili Yu and Wuming Zhang and Zhuoyang Zhang and Ury Zhilinsky},
      year={2026},
      eprint={2604.15483},
      archivePrefix={arXiv},
      primaryClass={cs.LG},
      url={https://arxiv.org/abs/2604.15483}, 
}

@misc{fang2026molmoact2actionreasoningmodels,
      title={MolmoAct2: Action Reasoning Models for Real-world Deployment}, 
      author={Haoquan Fang and Jiafei Duan and Donovan Clay and Sam Wang and Shuo Liu and Weikai Huang and Xiang Fan and Wei-Chuan Tsai and Shirui Chen and Yi Ru Wang and Shanli Xing and Jaemin Cho and Jae Sung Park and Ainaz Eftekhar and Peter Sushko and Karen Farley and Angad Wadhwa and Cole Harrison and Winson Han and Ying-Chun Lee and Eli VanderBilt and Rose Hendrix and Suveen Ellawela and Lucas Ngoo and Joyce Chai and Zhongzheng Ren and Ali Farhadi and Dieter Fox and Ranjay Krishna},
      year={2026},
      eprint={2605.02881},
      archivePrefix={arXiv},
      primaryClass={cs.RO},
      url={https://arxiv.org/abs/2605.02881}, 
}

@misc{nvidia2026cosmos3omnimodalworld,
      title={Cosmos 3: Omnimodal World Models for Physical AI}, 
      author={NVIDIA and : and Aditi and Niket Agarwal and Arslan Ali and Jon Allen and Martin Antolini and Adeline Aubame and Alisson Azzolini and Junjie Bai and Maciej Bala and Yogesh Balaji and Josh Bapst and Aarti Basant and Mukesh Beladiya and Mohammad Qazim Bhat and Zaid Pervaiz Bhat and Dan Blick and Vanni Brighella and Han Cai and Tiffany Cai and Eric Cameracci and Jiaxin Cao and Yulong Cao and Mark Carlson and Carlos Casanova and Ting-Yun Chang and Yan Chang and Yu-Wei Chao and Prithvijit Chattopadhyay and Roshan Chaudhari and Chieh-Yun Chen and Junyu Chen and Ke Chen and Qizhi Chen and Wenkai Chen and Xiaotong Chen and Yu Chen and An-Chieh Cheng and Click Cheng and Xiu Chia and Jeana Choi and Chaeyeon Chung and Wenyan Cong and Yin Cui and Magdalena Dadela and Nalin Dadhich and Wenliang Dai and Joyjit Daw and Alperen Degirmenci and Rodrigo Vieira Del Monte and Robert Denomme and Sameer Dharur and Marco Di Lucca and Ke Ding and Wenhao Ding and Yifan Ding and Yuzhu Dong and Nicole Drumheller and Yilun Du and Aigul Dzhumamuratova and Aleksandr Efitorov and Hamid Eghbalzadeh and Naomi Eigbe and Imad El Hanafi and Hassan Eslami and Benedikt Falk and Jiaojiao Fan and Jim Fan and Amol Fasale and Sergiy Fefilatyev and Liang Feng and Francesco Ferroni and Sanja Fidler and Xiao Fu and Vikram Fugro and Prashant Gaikwad and TJ Galda and Katelyn Gao and Yihuai Gao and Wenhang Ge and Sreyan Ghosh and Arushi Goel and Vivek Goel and Akash Gokul and Rama Govindaraju and Jinwei Gu and Miguel Guerrero and Elfie Guo and Aryaman Gupta and Siddharth Gururani and Hugo Hadfield and Song Han and Ankur Handa and Zekun Hao and Mohammad Harrim and Ali Hassani and Nathan Hayes-Roth and Yufan He and Chris Helvig and Cyrus Hogg and Madison Huang and Michael Huang and Sophia Huang and Yufan Huang and Jacob Huffman and DeLesley Hutchins and Suneel Indupuru and Boris Ivanovic and Arihant Jain and Joel Jang and Ryan Ji and Yanan Jian and Dongfu Jiang and Jingyi Jin and Atharva Joshi and Nikhilesh Joshi and Pranjali Joshi and Andy Ju and Jaehun Jung and Weiwei Kang and Scott Kassekert and Jan Kautz and Ashna Khetan and Julia Kiczka and Slawek Kierat and Gwanghyun Kim and Kuno Kim and Sunny Kim and Kezhi Kong and Xin Kong and Zhifeng Kong and Tomasz Kornuta and Egor Krivov and Hui Kuang and Saurav Kumar and Chia-Wen Kuo and George Kurian and Wojciech Kutak and JF Lafleche and Himangshu Lahkar and Omar Laymoun and Jayjun Lee and Sanggil Lee and Gabriele Leone and Boyi Li and Freya Li and Jiajun Li and Jinfeng Li and Ling Li and Pengcheng Li and Shangru Li and Tingle Li and Xiaolong Li and Xuan Li and Zhaoshuo Li and Zhiqi Li and Hao Liang and Maosheng Liao and Chen-Hsuan Lin and Tsung-Yi Lin and Ming-Yu Liu and Sifei Liu and Zihan Liu and Hai Loc Lu and Xiangyu Lu and Alice Luo and Ruipu Luo and Wenjie Luo and Jiangran Lyu and Martin Ding Ma and Nic Ma and Qianli Ma and Dawid Majchrowski and Louis Marcoux and Miguel Martin and Qing Miao and Ashkan Mirzaei and Shreyas Misra and Kaichun Mo and Durra Mohsin and Hyejin Moon and Pawel Morkisz and Saeid Motiian and Kirill Motkov and Seungjun Nah and Yashraj Narang and Deepak Narayanan and Thabang Ngazimbi and Julian Ouyang and Shubham Pachori and David Page and Yatian Pang and Sehwi Park and Mahesh Patekar and Mostofa Patwary and Marco Pavone and Trung Pham and Wei Ping and Soha Pouya and Shrimai Prabhumoye and Varun Praveen and Delin Qu and Hesam Rabeti and Morteza Ramezanali and Marilyn Reeb and Xuanchi Ren and Kristen Rumley and Wojciech Rymer and Jun Saito and Yeongho Seol and John Shao and Piyush Shekdar and Tianwei Shen and Humphrey Shi and Min Shi and Stella Shi and Kevin Shih and Mohammad Shoeybi and Mateusz Sieniawski and Shuran Song and Alexander Sotelo and Amir Sotoodeh and Sunil Srinivasa and Vignesh Srinivasakumar and Bartosz Stefaniak and Rahul Heinrich Steiger and Shangkun Sun and Jiaxiang Tang and Shitao Tang and Yangyang Tang and Yue Tang and Tolou Tavakkoli and Kayley Ting and Krzysztof Tomala and Wei-Cheng Tseng and Jibin Varghese and Sergei Vasilev and Thomas Volk and Raju Wagwani and Roger Waleffe and Andrew Z. Wang and Boxiang Wang and Haoxiang Wang and Qiao Wang and Shihao Wang and Shijie Wang and Ting-Chun Wang and Yan Wang and Yu Wang and Rohit Watve and David Wehr and Fangyin Wei and Xinshuo Weng and Jay Zhangjie Wu and Kedi Wu and Hongchi Xia and Summer Xiao and Tianjun Xiao and Kevin Xie and Daguang Xu and Jiashu Xu and Mengyao Xu and Ruqing Xu and Xingqian Xu and Yao Xu and Dinghao Yang and Dong Yang and Hans Yang and Xiaodong Yang and Xuning Yang and Yichu Yang and Yurong You and Zhiding Yu and Hao Yuan and Simon Yuen and Xiaohui Zeng and Pengcuo Zeren and Cindy Zha and Haotian Zhang and Jenny Zhang and Jing Zhang and Liangkai Zhang and Paris Zhang and Shun Zhang and Xuanmeng Zhang and Zhizheng Zhang and Ann Zhao and Yilin Zhao and Yuliya Zhautouskaya and Charles Zhou and Fengzhe Zhou and Shilin Zhu and Yuke Zhu and Dima Zhylko and Artur Zolkowski},
      year={2026},
      eprint={2606.02800},
      archivePrefix={arXiv},
      primaryClass={cs.CV},
      url={https://arxiv.org/abs/2606.02800}, 
}

@misc{wen2025dexvlavisionlanguagemodelplugin,
      title={DexVLA: Vision-Language Model with Plug-In Diffusion Expert for General Robot Control}, 
      author={Junjie Wen and Yichen Zhu and Jinming Li and Zhibin Tang and Chaomin Shen and Feifei Feng},
      year={2025},
      eprint={2502.05855},
      archivePrefix={arXiv},
      primaryClass={cs.RO},
      url={https://arxiv.org/abs/2502.05855}, 
}

@misc{kumar2022finetuningdistortpretrainedfeatures,
      title={Fine-Tuning can Distort Pretrained Features and Underperform Out-of-Distribution}, 
      author={Ananya Kumar and Aditi Raghunathan and Robbie Jones and Tengyu Ma and Percy Liang},
      year={2022},
      eprint={2202.10054},
      archivePrefix={arXiv},
      primaryClass={cs.LG},
      url={https://arxiv.org/abs/2202.10054}, 
}

@misc{hwang2025emmaendtoendmultimodalmodel,
      title={EMMA: End-to-End Multimodal Model for Autonomous Driving}, 
      author={Jyh-Jing Hwang and Runsheng Xu and Hubert Lin and Wei-Chih Hung and Jingwei Ji and Kristy Choi and Di Huang and Tong He and Paul Covington and Benjamin Sapp and Yin Zhou and James Guo and Dragomir Anguelov and Mingxing Tan},
      year={2025},
      eprint={2410.23262},
      archivePrefix={arXiv},
      primaryClass={cs.CV},
      url={https://arxiv.org/abs/2410.23262}, 
}

@misc{zhou2025opendrivevlaendtoendautonomousdriving,
      title={OpenDriveVLA: Towards End-to-end Autonomous Driving with Large Vision Language Action Model}, 
      author={Xingcheng Zhou and Xuyuan Han and Feng Yang and Yunpu Ma and Volker Tresp and Alois Knoll},
      year={2025},
      eprint={2503.23463},
      archivePrefix={arXiv},
      primaryClass={cs.CV},
      url={https://arxiv.org/abs/2503.23463}, 
}

@misc{zhou2025autovlavisionlanguageactionmodelendtoend,
      title={AutoVLA: A Vision-Language-Action Model for End-to-End Autonomous Driving with Adaptive Reasoning and Reinforcement Fine-Tuning}, 
      author={Zewei Zhou and Tianhui Cai and Seth Z. Zhao and Yun Zhang and Zhiyu Huang and Bolei Zhou and Jiaqi Ma},
      year={2025},
      eprint={2506.13757},
      archivePrefix={arXiv},
      primaryClass={cs.CV},
      url={https://arxiv.org/abs/2506.13757}, 
}

@misc{han2026physicsmultimodalpretrainingknowledge,
      title={Towards Physics of Multimodal Pretraining: Knowledge Flow, Modality Synergy, Early Unification, and Recipes}, 
      author={Junlin Han and Shengbang Tong and David Fan and Minghao Chen and Philip Torr and Filippos Kokkinos and Mike Lewis},
      year={2026},
      eprint={2608.05000},
      archivePrefix={arXiv},
      primaryClass={cs.CV},
      url={https://arxiv.org/abs/2608.05000}, 
}

@misc{yuan2026qwenrobotmaniptechnicalreportalignment,
      title={Qwen-RobotManip Technical Report: Alignment Unlocks Scale for Robotic Manipulation Foundation Models}, 
      author={Haoqi Yuan and Zhixuan Liang and Anzhe Chen and Ye Wang and Haoyang Li and Pei Lin and Yiyang Huang and Zixing Lei and Tong Zhang and Jiazhao Zhang and Jie Zhang and Jingyang Fan and Gengze Zhou and Qihang Peng and Chenxu Lv and Xiaoyue Chen and An Yang and Fei Huang and Junyang Lin and Dayiheng Liu and Jingren Zhou and Chenfei Wu and Xiong-Hui Chen},
      year={2026},
      eprint={2606.17846},
      archivePrefix={arXiv},
      primaryClass={cs.RO},
      url={https://arxiv.org/abs/2606.17846}, 
}

@misc{shukor2025smolvlavisionlanguageactionmodelaffordable,
      title={SmolVLA: A Vision-Language-Action Model for Affordable and Efficient Robotics}, 
      author={Mustafa Shukor and Dana Aubakirova and Francesco Capuano and Pepijn Kooijmans and Steven Palma and Adil Zouitine and Michel Aractingi and Caroline Pascal and Martino Russi and Andres Marafioti and Simon Alibert and Matthieu Cord and Thomas Wolf and Remi Cadene},
      year={2025},
      eprint={2506.01844},
      archivePrefix={arXiv},
      primaryClass={cs.LG},
      url={https://arxiv.org/abs/2506.01844}, 
}

@article{jain2025polaris,
  title={Polaris: Scalable real-to-sim evaluations for generalist robot policies},
  author={Jain, Arhan and Zhang, Mingtong and Arora, Kanav and Chen, William and Torne, Marcel and Irshad, Muhammad Zubair and Zakharov, Sergey and Wang, Yue and Levine, Sergey and Finn, Chelsea and others},
  journal={arXiv preprint arXiv:2512.16881},
  year={2025}
}

@misc{nvidia_physicalai_av_2026,
    author       = {{NVIDIA Corporation}},
    title        = {{PhysicalAI-Autonomous-Vehicles}},
    year         = {2026},
    howpublished = {Hugging Face dataset},
    url          = {https://huggingface.co/datasets/nvidia/PhysicalAI-Autonomous-Vehicles},
    note         = {Version 26.09. Accessed September 23, 2026}
  }

@software{alpasim_2025,
  author       = {
    NVIDIA and
    Yulong Cao and
    Riccardo de Lutio and
    Sanja Fidler and
    Guillermo Garcia Cobo and
    Zan Gojcic and
    Maximilian Igl and
    Boris Ivanovic and
    Peter Karkus and
    Janick Martinez Esturo and
    Marco Pavone and
    Aaron Smith and
    Ellie Tanimura and
    Michal Tyszkiewicz and
    Michael Watson and
    Qi Wu and
    Le Zhang
  },
  title        = {AlpaSim: A Modular, Lightweight, and Data-Driven Research Simulator for Autonomous Driving},
  year         = {2025},
  month        = {October},
  url          = {https://github.com/NVlabs/alpasim},
}

@misc{beyer2024paligemmaversatile3bvlm,
      title={PaliGemma: A versatile 3B VLM for transfer}, 
      author={Lucas Beyer and Andreas Steiner and André Susano Pinto and Alexander Kolesnikov and Xiao Wang and Daniel Salz and Maxim Neumann and Ibrahim Alabdulmohsin and Michael Tschannen and Emanuele Bugliarello and Thomas Unterthiner and Daniel Keysers and Skanda Koppula and Fangyu Liu and Adam Grycner and Alexey Gritsenko and Neil Houlsby and Manoj Kumar and Keran Rong and Julian Eisenschlos and Rishabh Kabra and Matthias Bauer and Matko Bošnjak and Xi Chen and Matthias Minderer and Paul Voigtlaender and Ioana Bica and Ivana Balazevic and Joan Puigcerver and Pinelopi Papalampidi and Olivier Henaff and Xi Xiong and Radu Soricut and Jeremiah Harmsen and Xiaohua Zhai},
      year={2024},
      eprint={2407.07726},
      archivePrefix={arXiv},
      primaryClass={cs.CV},
      url={https://arxiv.org/abs/2407.07726}, 
}

@misc{goyal2017makingvvqamatter,
      title={Making the V in VQA Matter: Elevating the Role of Image Understanding in Visual Question Answering}, 
      author={Yash Goyal and Tejas Khot and Douglas Summers-Stay and Dhruv Batra and Devi Parikh},
      year={2017},
      eprint={1612.00837},
      archivePrefix={arXiv},
      primaryClass={cs.CV},
      url={https://arxiv.org/abs/1612.00837}, 
}

@misc{chen2015microsoftcococaptionsdata,
      title={Microsoft COCO Captions: Data Collection and Evaluation Server}, 
      author={Xinlei Chen and Hao Fang and Tsung-Yi Lin and Ramakrishna Vedantam and Saurabh Gupta and Piotr Dollar and C. Lawrence Zitnick},
      year={2015},
      eprint={1504.00325},
      archivePrefix={arXiv},
      primaryClass={cs.CV},
      url={https://arxiv.org/abs/1504.00325}, 
}

@misc{lin2015microsoftcococommonobjects,
      title={Microsoft COCO: Common Objects in Context}, 
      author={Tsung-Yi Lin and Michael Maire and Serge Belongie and Lubomir Bourdev and Ross Girshick and James Hays and Pietro Perona and Deva Ramanan and C. Lawrence Zitnick and Piotr Dollár},
      year={2015},
      eprint={1405.0312},
      archivePrefix={arXiv},
      primaryClass={cs.CV},
      url={https://arxiv.org/abs/1405.0312}, 
}

@misc{krishna2016visualgenomeconnectinglanguage,
      title={Visual Genome: Connecting Language and Vision Using Crowdsourced Dense Image Annotations}, 
      author={Ranjay Krishna and Yuke Zhu and Oliver Groth and Justin Johnson and Kenji Hata and Joshua Kravitz and Stephanie Chen and Yannis Kalantidis and Li-Jia Li and David A. Shamma and Michael S. Bernstein and Fei-Fei Li},
      year={2016},
      eprint={1602.07332},
      archivePrefix={arXiv},
      primaryClass={cs.CV},
      url={https://arxiv.org/abs/1602.07332}, 
}

@misc{khazatsky2025droidlargescaleinthewildrobot,
      title={DROID: A Large-Scale In-The-Wild Robot Manipulation Dataset}, 
      author={Alexander Khazatsky and Karl Pertsch and Suraj Nair and Ashwin Balakrishna and Sudeep Dasari and Siddharth Karamcheti and Soroush Nasiriany and Mohan Kumar Srirama and Lawrence Yunliang Chen and Kirsty Ellis and Peter David Fagan and Joey Hejna and Masha Itkina and Marion Lepert and Yecheng Jason Ma and Patrick Tree Miller and Jimmy Wu and Suneel Belkhale and Shivin Dass and Huy Ha and Arhan Jain and Abraham Lee and Youngwoon Lee and Marius Memmel and Sungjae Park and Ilija Radosavovic and Kaiyuan Wang and Albert Zhan and Kevin Black and Cheng Chi and Kyle Beltran Hatch and Shan Lin and Jingpei Lu and Jean Mercat and Abdul Rehman and Pannag R Sanketi and Archit Sharma and Cody Simpson and Quan Vuong and Homer Rich Walke and Blake Wulfe and Ted Xiao and Jonathan Heewon Yang and Arefeh Yavary and Tony Z. Zhao and Christopher Agia and Rohan Baijal and Mateo Guaman Castro and Daphne Chen and Qiuyu Chen and Trinity Chung and Jaimyn Drake and Ethan Paul Foster and Jensen Gao and Vitor Guizilini and David Antonio Herrera and Minho Heo and Kyle Hsu and Jiaheng Hu and Muhammad Zubair Irshad and Donovon Jackson and Charlotte Le and Yunshuang Li and Kevin Lin and Roy Lin and Zehan Ma and Abhiram Maddukuri and Suvir Mirchandani and Daniel Morton and Tony Nguyen and Abigail O'Neill and Rosario Scalise and Derick Seale and Victor Son and Stephen Tian and Emi Tran and Andrew E. Wang and Yilin Wu and Annie Xie and Jingyun Yang and Patrick Yin and Yunchu Zhang and Osbert Bastani and Glen Berseth and Jeannette Bohg and Ken Goldberg and Abhinav Gupta and Abhishek Gupta and Dinesh Jayaraman and Joseph J Lim and Jitendra Malik and Roberto Martín-Martín and Subramanian Ramamoorthy and Dorsa Sadigh and Shuran Song and Jiajun Wu and Michael C. Yip and Yuke Zhu and Thomas Kollar and Sergey Levine and Chelsea Finn},
      year={2025},
      eprint={2403.12945},
      archivePrefix={arXiv},
      primaryClass={cs.RO},
      url={https://arxiv.org/abs/2403.12945}, 
}

@misc{kwok2026scalingverificationeffectivescaling,
      title={Scaling Verification Can Be More Effective than Scaling Policy Learning for Vision-Language-Action Alignment}, 
      author={Jacky Kwok and Xilun Zhang and Mengdi Xu and Yuejiang Liu and Azalia Mirhoseini and Chelsea Finn and Marco Pavone},
      year={2026},
      eprint={2602.12281},
      archivePrefix={arXiv},
      primaryClass={cs.RO},
      url={https://arxiv.org/abs/2602.12281}, 
}

@misc{zhou2026qwendrive10initialstepvisionlanguage,
      title={Qwen-Drive-1.0: An Initial Step towards a Vision-Language Foundation Model for Autonomous Driving}, 
      author={Xin Zhou and Zongchuang Zhao and Zhibo Yang and Mingsheng Li and Humen Zhong and Shuai Bai and Du Chu and Ruizhe Chen and Zhaohai Li and Jun Tang and Qiuyue Wang and Mingkun Yang and Jiazhao Zhang and Dayiheng Liu and Dingkang Liang and Xiang Bai},
      year={2026},
      eprint={2609.00111},
      archivePrefix={arXiv},
      primaryClass={cs.CV},
      url={https://arxiv.org/abs/2609.00111}, 
}

@misc{zhou2026spanvlaefficientactionbridging,
      title={SpanVLA: Efficient Action Bridging and Learning from Negative-Recovery Samples for Vision-Language-Action Model}, 
      author={Zewei Zhou and Ruining Yang and Xuewei and Qi and Yiluan Guo and Sherry X. Chen and Tao Feng and Kateryna Pistunova and Yishan Shen and Lili Su and Jiaqi Ma},
      year={2026},
      eprint={2604.19710},
      archivePrefix={arXiv},
      primaryClass={cs.CV},
      url={https://arxiv.org/abs/2604.19710}, 
}

@misc{li2026unidrivevlaunifyingunderstandingperception,
      title={UniDriveVLA: Unifying Understanding, Perception, and Action Planning for Autonomous Driving}, 
      author={Yongkang Li and Lijun Zhou and Sixu Yan and Bencheng Liao and Tianyi Yan and Kaixin Xiong and Long Chen and Hongwei Xie and Bing Wang and Guang Chen and Hangjun Ye and Wenyu Liu and Haiyang Sun and Xinggang Wang},
      year={2026},
      eprint={2604.02190},
      archivePrefix={arXiv},
      primaryClass={cs.CV},
      url={https://arxiv.org/abs/2604.02190}, 
}

@misc{yang2025qwen3technicalreport,
      title={Qwen3 Technical Report}, 
      author={An Yang and Anfeng Li and Baosong Yang and Beichen Zhang and Binyuan Hui and Bo Zheng and Bowen Yu and Chang Gao and Chengen Huang and Chenxu Lv and Chujie Zheng and Dayiheng Liu and Fan Zhou and Fei Huang and Feng Hu and Hao Ge and Haoran Wei and Huan Lin and Jialong Tang and Jian Yang and Jianhong Tu and Jianwei Zhang and Jianxin Yang and Jiaxi Yang and Jing Zhou and Jingren Zhou and Junyang Lin and Kai Dang and Keqin Bao and Kexin Yang and Le Yu and Lianghao Deng and Mei Li and Mingfeng Xue and Mingze Li and Pei Zhang and Peng Wang and Qin Zhu and Rui Men and Ruize Gao and Shixuan Liu and Shuang Luo and Tianhao Li and Tianyi Tang and Wenbiao Yin and Xingzhang Ren and Xinyu Wang and Xinyu Zhang and Xuancheng Ren and Yang Fan and Yang Su and Yichang Zhang and Yinger Zhang and Yu Wan and Yuqiong Liu and Zekun Wang and Zeyu Cui and Zhenru Zhang and Zhipeng Zhou and Zihan Qiu},
      year={2025},
      eprint={2505.09388},
      archivePrefix={arXiv},
      primaryClass={cs.CL},
      url={https://arxiv.org/abs/2505.09388}, 
}

@misc{marcu2024lingoqavisualquestionanswering,
      title={LingoQA: Visual Question Answering for Autonomous Driving}, 
      author={Ana-Maria Marcu and Long Chen and Jan Hünermann and Alice Karnsund and Benoit Hanotte and Prajwal Chidananda and Saurabh Nair and Vijay Badrinarayanan and Alex Kendall and Jamie Shotton and Elahe Arani and Oleg Sinavski},
      year={2024},
      eprint={2312.14115},
      archivePrefix={arXiv},
      primaryClass={cs.RO},
      url={https://arxiv.org/abs/2312.14115}, 
}

\clearpage
\appendix
\raggedbottom
\section{Implementation and Training Details}
\label{app:method_details}
\label{app:training_details}

This section reports the architecture, data mixtures, optimizer settings, and training schedules for the driving and manipulation experiments.

\subsection{Driving model and action representation}
\label{app:driving_architecture}

The driving policy uses a Qwen3-VL-2B-Instruct \citep{yang2025qwen3technicalreport} backbone and a non-causal flow-matching action expert. At each layer, the expert attends to the K/V representations of the corresponding VLM layer. Insulation detaches these representations before the expert reads them, leaving their forward values unchanged. Under coupling, the K/V representations remain attached and the flow loss can update the VLM.

The text tower and action expert use 128-dimensional attention heads. The vision tower outputs 2,048-dimensional features from $16\times16$ patches with spatial merge size two. The action expert uses a two-layer waypoint input projection with width 512 and 20 Fourier features, followed by a linear output projection. The VLM context length is 16,384 tokens.

The expert represents the 6.4\,s trajectory as 64 acceleration--curvature
control pairs sampled at 10\,Hz and converts them to pose waypoints using
discretized unicycle dynamics. The VLM receives quantized ego-history
tokens and, when action-token supervision is enabled, predicts a
128-token discretization of the same future control sequence.

\begin{table}[!htbp]

    \centering
    \small
    \captionsetup{width=\linewidth}
    \setlength{\tabcolsep}{5pt}
    \renewcommand{\arraystretch}{1.12}
    \caption{\textbf{Driving optimization settings.} Shared across conditions within each training history.}
    \label{tab:app_driving_optimization}
    \begin{tabular}{@{}lr@{}}
        \toprule
        \textbf{Setting} & \textbf{Value} \\
        \midrule
        Optimizer & Fused AdamW \\
        $(\beta_1,\beta_2)$ & $(0.9,0.95)$ \\
        Adam $\epsilon$ & $10^{-8}$ \\
        Weight decay & $0.1$ \\
        Vision LR & $10^{-5}$ \\
        VLM text/head LR & $10^{-4}$ \\
        Expert LR & $4\times10^{-4}$ \\
        LR schedule & Cosine, 400k steps \\
        Minimum LR factor & $0.01$ \\
        Global batch & 128 \\
        Gradient clipping & $1.0$ \\
        Model precision & Bfloat16 \\
        Optimizer master & Float32 \\
        \bottomrule
    \end{tabular}
\end{table}

When future-action CE is disabled, the 128 future value-token targets are masked, but their teacher-forced input tokens remain in the sequence. Past-trajectory values and the future-trajectory boundary markers remain supervised. The CE denominator is the number of supervised positions before masking, so removing future-action CE does not increase the weight of the retained autoregressive targets. The expert-facing K/V cache ends immediately after the \texttt{<future\_start>} marker: it contains neither future value-token states nor the \texttt{<future\_end>} state. Consequently, the expert cannot read the teacher-forced future values.

More precisely, let $\mathcal{F}$ contain future-trajectory value-token target positions, let $\mathcal{O}$ contain the other supervised positions, and let $\ell_i$ be next-token CE at position $i$. For $p=1$ when action-token supervision is on and $p=0$ when it is off, the implemented loss is
\begin{equation}
    \mathcal{L}_{\mathrm{CE}}^{(p)}=
    \frac{\sum_{i\in\mathcal{O}}\ell_i+p\sum_{i\in\mathcal{F}}\ell_i}{|\mathcal{O}|+|\mathcal{F}|}.
    \label{eq:app_driving_ce}
\end{equation}
The non-action CE contribution therefore has the same scale under both settings. The flow and CE objectives each have unit weight on trajectory-bearing batches.

\subsection{Driving flow matching and optimization}

The driving flow path is $a_t=t a+(1-t)\epsilon$, with target vector field $a-\epsilon$ and $\epsilon\sim\mathcal{N}(0,I)$. We draw $b\sim\operatorname{Beta}(1.5,1.0)$ and set $t=0.999(1-b)$. At inference, the expert starts from Gaussian noise and uses 10 Euler steps. Classifier-free guidance is disabled.

Continous action-prediction batches have sampling probability $48.5\%$, and vision--language batches have probability $\approx51.5\%$. Training examples with one, two, four, or seven cameras are sampled with relative weights $1{:}2{:}3{:}4$; in evaluation, we always use four RGB images. The expert receives a flow update only on action-bearing batches, whereas the VLM receives an autoregressive update on every joint-training step. Each training run used 64 NVIDIA H100 GPUs and consumed approximately 15,000 GPU-hours on average. Table~\ref{tab:app_driving_optimization} reports the common optimization settings.

The concurrent conditions start from identical VLM and randomly initialized expert weights. Expert-First conditions share a 100k-step expert checkpoint trained on trajectory data with the VLM frozen, then receive 200k joint-training steps. VLM-First conditions train the VLM alone for 100k steps, either with or without future-action CE, before attaching experts with a shared fresh initialization for 200k joint-training steps. The VLM-First action-token-off warm-up uses the vision--language mixture without action-bearing batches. Comparisons within each history have matched training schedules and update opportunities.

Table~\ref{tab:update_budget} gives the expected component-level gradient exposure at the 300k-total-step endpoints and the intermediate VLM-First checkpoint. The Expert-First warm-up uses only trajectory data, so every warm-up step updates the expert. In VLM-First training, action-token supervision also changes the first-stage data mixture: the tokens-off backbone is trained on vision--language data alone, while the tokens-on backbone sees trajectory-prediction data. The four conditions share the same joint-training mixture after the expert is introduced.

\begin{table}[htbp]
    \centering
    \small
    \setlength{\tabcolsep}{5pt}
    \renewcommand{\arraystretch}{1.16}
    \caption{\textbf{Expected component updates.} Endpoints have 300k total steps}
    \label{tab:update_budget}
    \label{tab:app_update_budgets}
    \begin{tabular*}{\linewidth}{@{\extracolsep{\fill}}llrr@{}}
        \toprule
        \textbf{History} & \textbf{Schedule} & \textbf{VLM updates} & \textbf{Expert updates} \\
        \midrule
        Concurrent & 300k joint steps & 300k & $\approx145.5$k \\
        Expert-First & 100k expert-only + 200k joint & 200k & $\approx197$k \\
        VLM-First & 100k VLM-only + 200k joint & 300k & $\approx97$k \\
        \bottomrule
    \end{tabular*}
\end{table}

\subsection{Manipulation implementation and training}
\label{app:manipulation_setup}

We compare two manipulation model families: a $\pi_{0.5}$ re-implementation and MolmoAct2. Within each family, the four conditions vary FAST supervision and expert-to-VLM gradient flow while sharing the data mixture, initialization procedure, and optimization schedule. The architectures and training settings differ across families, as described below.

\paragraph{$\pi_{0.5}$ implementation.}
The $\pi_{0.5}$ manipulation policy pairs a PaliGemma VLM with a flow-matching action expert (Table~\ref{tab:app_manipulation_architecture}). Observations contain an external RGB image and a left-wrist RGB image, each resized to $224\times224$, together with seven robot joint values and one gripper value. The state vector is padded to 32 dimensions. The expert predicts a 16-step action chunk in a 32-dimensional padded space; the first eight output dimensions encode seven joint values and one gripper value. During training, joint positions are expressed as deltas, while the gripper value remains absolute.

\begin{table}[htbp]
    \centering
    \small
    \setlength{\tabcolsep}{6pt}
    \renewcommand{\arraystretch}{1.16}
    \caption{\textbf{Manipulation model dimensions.} Attention heads are query/key-value counts.}
    \label{tab:app_manipulation_architecture}
    \begin{tabular*}{\linewidth}{@{\extracolsep{\fill}}llrrrr@{}}
        \toprule
        \textbf{Component} & \textbf{Backbone} & \textbf{Layers} & \textbf{Width} & \textbf{MLP} & \textbf{Heads (Q/KV)} \\
        \midrule
        \multicolumn{6}{@{}l}{$\pi_{0.5}$} \\
        VLM Backbone & Gemma 2B & 18 & 2,048 & 16,384 & 8/1 \\
        Action expert & Gemma 300M expert (random) & 18 & 1,024 & 4,096 & 8/1 \\
        \midrule
        \multicolumn{6}{@{}l}{MolmoAct2} \\
        VLM Backbone & Molmo2-ER & 36 & 2,560 & 19,456 & 32/8 \\
        Action expert & MolmoAct2 expert (random) & 36 & 768 & 3,072 & 8/8 \\
        \bottomrule
    \end{tabular*}
\end{table}

As in driving, insulation stops the expert's gradient at the backbone K/V interface while preserving the forward representations. In FAST-supervised robot examples, the autoregressive branch predicts a FAST action postfix. Without FAST supervision, the postfix is absent and robot examples contribute no action-token CE; vision--language examples retain their language-answer CE in every condition. The expert is prevented from attending to FAST-token keys. This implementation differs from the driving intervention, where masked future-action value tokens remain in the training sequence. In the $\pi_{0.5}$ implementation, CE loss is normalized by the number of active target positions per example, clipped to at least one, and the flow loss is applied only to robot-action examples.

Each $\pi_{0.5}$ training batch draws a binomially distributed number of vision--language examples with probability one half; the remaining examples come from the robot stream. The streams therefore contribute equally in expectation, although individual batches need not contain exactly half of each type. The robot stream samples 90\% DROID data and 10\% PolaRiS data. Within the vision--language stream, task weights are 35\% VQAv2 question answering, 20\% COCO captioning, 20\% COCO localization, 15\% Visual Genome question answering, and 10\% Visual Genome region tasks. A deterministic 2\% hash split is held out for validation; robot examples are split by episode, and COCO examples use image identifiers where available to prevent cross-task image overlap. Table~\ref{tab:app_manipulation_optimization} records the $\pi_{0.5}$ manipulation training settings. Inference uses 10 Euler steps from Gaussian noise.

\begin{table}[!htbp]
    \centering
    \small
    \setlength{\tabcolsep}{5pt}
    \renewcommand{\arraystretch}{1.12}
    \caption{\textbf{$\pi_{0.5}$ manipulation optimization settings.} Shared across conditions through 400k steps.}
    \label{tab:app_manipulation_optimization}
    \begin{tabular}{@{}lr@{}}
        \toprule
        \textbf{Setting} & \textbf{Value} \\
        \midrule
        Optimizer & AdamW \\
        $(\beta_1,\beta_2)$ & $(0.9,0.95)$ \\
        Adam $\epsilon$ & $10^{-8}$ \\
        Weight decay & $10^{-10}$ \\
        Peak/terminal LR & $5\times10^{-5}$ \\
        Warm-up & 1,000 steps \\
        Global batch & 128 \\
        Gradient clipping & $1.0$ \\
        EMA decay & $0.99$ \\
        \bottomrule
    \end{tabular}
\end{table}

\paragraph{MolmoAct2 implementation.}
We use the released MolmoAct2 implementation \citep{fang2026molmoact2actionreasoningmodels}, initializing each condition from the official \texttt{allenai/Molmo2-ER} checkpoint with a fresh, randomly initialized action expert. The expert conditions on the VLM's layerwise K/V representations. Insulation stops the expert's gradient at this interface while preserving the forward representations; without insulation, the flow-loss gradient reaches the VLM. In FAST-supervised robot batches, training jointly optimizes FAST-token CE and continuous flow matching using \texttt{allenai/MolmoAct2-FAST-Tokenizer}. Without FAST supervision, FAST action targets are absent and robot batches retain only continuous flow matching. Separate multimodal batches retain their language objective and contribute no expert loss. We retain the implementation's token-loss weighting and global-batch normalization rather than the per-example CE normalization used for $\pi_{0.5}$.

All four conditions train for 300k steps with the action expert present and trainable from the outset. This differs from the released staged recipe, which first trains discrete action prediction and subsequently adds the continuous expert, but keeps relative training and compute comparable. Depth reasoning and depth-token supervision are disabled. The VLM, vision encoder, embedder, and action expert remain trainable in every condition.

\paragraph{MolmoAct2 data mixture.}
Each MolmoAct2 optimizer step draws a robot or multimodal batch with probabilities 90\% and 10\%, respectively. Within the robot stream, sampling weights are 30\% YAM, 30\% SO100/SO101, 30\% DROID, and 2.5\% each for MolmoAct, BC-Z, RT-1, and Bridge. Our public reconstruction includes all 737 YAM repositories, 1,202 of the 1,220 SO100/SO101 repositories in the released configuration, and all six consolidated repositories covering the remaining robot groups. The 18 unavailable SO repositories account for approximately 0.398\% of the original overall loader-selection mass; their weight is redistributed within the SO group.

Within the multimodal stream, sampling weights are 46\% Molmo2, 46\% Molmo2-ER, and 8\% Tulu-3. We use the publicly available sources and redistribute the weights of unavailable sources within each group. All four conditions use the same data mixture. We retain the released embodiment-specific robot processing and quantile normalization. For SO datasets with missing normalization statistics, we estimate them from the converted data.

\paragraph{MolmoAct2 optimization.}
Training uses global batch 128 and bfloat16 mixed precision on 64 H100 GPUs. We use AdamW with $(\beta_1,\beta_2)=(0.9,0.95)$, $\epsilon=10^{-6}$, and the learning rates reported in Table~\ref{tab:app_molmoact2_optimization}. After 200 warm-up steps, learning rates decay with a cosine schedule to 10\% of peak at step 200k and remain fixed through step 300k. Gradient clipping uses threshold $1.0$.

\begin{table}[!htbp]
    \centering
    \small
    \setlength{\tabcolsep}{6pt}
    \renewcommand{\arraystretch}{1.12}
    \caption{\textbf{MolmoAct2 learning rates.} Shared across conditions and used for both training and adaptation.}
    \label{tab:app_molmoact2_optimization}
    \begin{tabular}{@{}lr@{}}
        \toprule
        \textbf{Component} & \textbf{Peak LR} \\
        \midrule
        Language model & $1 \times10^{-5}$ \\
        Vision encoder & $5\times10^{-6}$ \\
        Connector & $5\times10^{-6}$ \\
        Action expert & $5\times10^{-5}$ \\
        \bottomrule
    \end{tabular}
\end{table}

\paragraph{MolmoAct2 adaptation and inference.}
Each final 300k-step checkpoint receives 1,000 adaptation steps on 90\% DROID and 10\% PolaRiS training data. Adaptation uses global batch 128 and the learning rates reported in Table~\ref{tab:app_molmoact2_optimization}, with 200 warm-up steps followed by cosine decay to 10\% of peak. Each condition retains its FAST supervision and insulation settings.

Observations contain an exterior RGB image, a left-wrist RGB image, seven robot joint values, one gripper value, and the task instruction. Images use MolmoAct2's native processing at $378\times378$, and states use its discrete encoding and saved normalization statistics. The expert predicts 15-step action chunks with eight active dimensions in a 32-dimensional padded space. Joint positions and the gripper value are absolute, and the gripper is not quantile-normalized. Inference uses 10 flow-integration steps.

We also adapt the released \texttt{allenai/MolmoAct2} checkpoint for 1,000 steps on the same DROID/PolaRiS mixture. These reference models use FAST supervision with insulation off during adaptation and are evaluated separately from the four factorial conditions.

\section{Evaluation and Statistical Details}
\label{app:evaluation_details}

\subsection{Open-loop driving}

The Public Golden evaluation dataset contains 1,434 examples; the out-of-distribution (OOD) evaluation contains 273 examples. Both evaluations use four cameras, four temporal frames per camera, and a 16-waypoint ego-trajectory history. The policy generates a chain of thought before sampling six continuous future trajectories. Chain-of-thought generation uses temperature $0.6$, top-$p$ $0.98$, and a maximum of 256 generated tokens, which we found realized the best performance. All driving results in this paper use this inference mode and the same dataset and prompt configuration across training histories.

For each candidate trajectory, ADE is the mean Euclidean distance in the ground plane between predicted and ground-truth waypoints over the 64-step horizon, measured in meters. Minimum ADE selects the lowest-ADE candidate among the six using ground truth.
For open-loop bootstrap confidence intervals, we resample base-clip clusters with replacement and retain every example belonging to each sampled clip. We apply the same sampled clips, with the same multiplicities, to all four conditions before recomputing condition means and paired differences. We use 10,000 such resamples and report the 2.5th and 97.5th percentiles of each bootstrap distribution as the 95\% interval. 

\FloatBarrier
\section{Additional Manipulation Results}
\label{app:additional_manipulation_results}

We evaluate all four conditions within each model family in three PolaRiS DROID environments \citep{jain2025polaris}: BlockStackKitchen, FoodBussing, and PanClean. The $\pi_{0.5}$ policies are evaluated after 400k training steps, and the MolmoAct2 policies after 300k training steps followed by 1,000 adaptation steps. PolaRiS places reconstructions of real scenes in a physics simulator, allowing policies to be compared from matched initial conditions. Each environment uses one instruction, held fixed across conditions:

\begin{itemize}
    \item \textbf{BlockStackKitchen:} ``Place and stack the blocks on top of the green tray''
    \item \textbf{FoodBussing:} ``Put all the foods in the bowl''
    \item \textbf{PanClean:} ``Use the yellow sponge to scrub the blue handle frying pan''
\end{itemize}

\begin{figure}[!htbp]
    \centering
    \includegraphics[width=\linewidth]{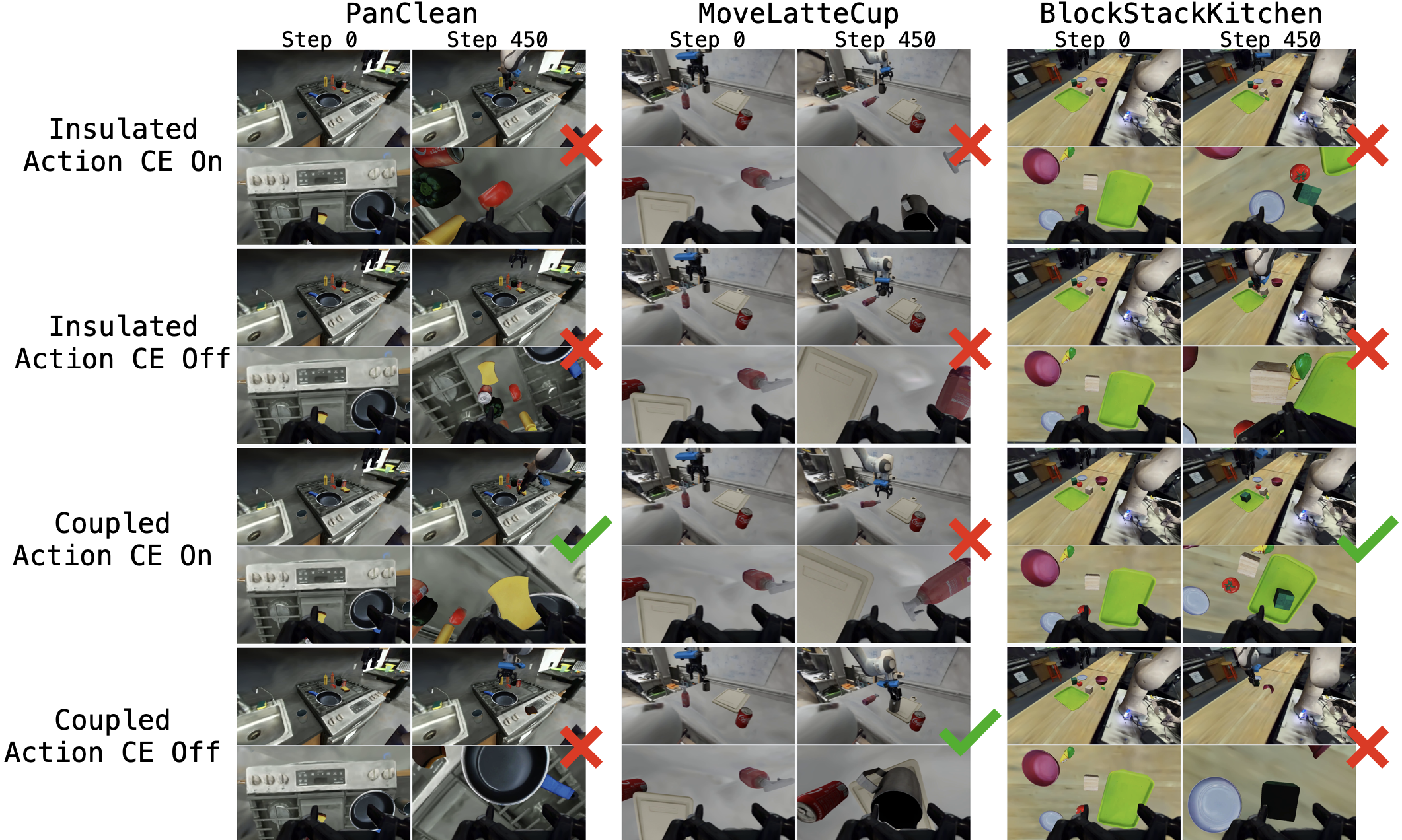}
    \caption{\textbf{Closed-loop manipulation examples.} Rollouts of the four concurrently trained policies in \texttt{PanClean} (left), \texttt{MoveLatteCup} (center), and \texttt{BlockStackKitchen} (right). \texttt{MoveLatteCup} is an additional qualitative example and is excluded from the three-task averages. Check marks indicate task success and crosses indicate failure. Among these examples, only the coupled policies succeed.}
    \label{fig:manipulation_concurrent_qualitative}
\end{figure}

\paragraph{Evaluation protocol.} Each condition is evaluated on the same initial conditions in every environment, which vary per episode, pairing the 50 episodes by scene across conditions. Episodes run for the default 450-step horizon without stopping early on success. The policies receive exterior and wrist RGB images and predict seven joint positions and a gripper command. For $\pi_{0.5}$, images are resized with padding to $224\times224$; MolmoAct2 uses its native image processing at $378\times378$. Importantly, we match the training-time inputs in the simulator to prevent distribution shift and give an accurate assessment of policy performance. Actions are executed open-loop in eight-step chunks. We use one training seed per condition.

\paragraph{Scoring.} Normalized task progress is the fraction of an environment's rubric criteria satisfied. Each criterion latches once satisfied; criteria with dependencies count only after the required criteria have been satisfied. Progress therefore cannot decrease, and success requires all criteria. Table~\ref{tab:app_polaris_rubrics} specifies the conditions and dependencies. For PanClean, the rubric scores placement of the sponge on the pan; it does not separately score a scrubbing motion. Figure~\ref{fig:manipulation_concurrent_qualitative} showcases qualitative demonstrations of the rollouts in simulation.

\begin{table}[!htbp]
    \centering
    \footnotesize
    \renewcommand{\arraystretch}{0.9}
    \setlength{\tabcolsep}{4pt}
    \caption{\textbf{PolaRiS rubric criteria.} Overlap is measured between horizontal bounding boxes. Placement requires the gripper to be open.}
    \label{tab:app_polaris_rubrics}
    \begin{tabular}{@{}cllc@{}}
        \toprule
        \# & Criterion & Condition & Requires \\
        \midrule
        \multicolumn{4}{@{}l}{\textbf{PanClean}} \\
        1 & Reach sponge & End effector within $0.2\,\mathrm{m}$ of sponge & --- \\
        2 & Lift sponge & Sponge height $>0.09\,\mathrm{m}$ & 1 \\
        3 & Sponge on pan & At least 80\% overlap with pan; gripper open & 2 \\
        \midrule
        \multicolumn{4}{@{}l}{\textbf{FoodBussing}} \\
        1 & Reach ice cream & End effector within $0.2\,\mathrm{m}$ & --- \\
        2 & Reach grapes & End effector within $0.2\,\mathrm{m}$ & --- \\
        3 & Lift ice cream & More than $0.06\,\mathrm{m}$ above rest height & 1 \\
        4 & Lift grapes & More than $0.06\,\mathrm{m}$ above rest height & 2 \\
        5 & Ice cream in bowl & At least 80\% overlap with bowl; gripper open & 3 \\
        6 & Grapes in bowl & At least 80\% overlap with bowl; gripper open & 4 \\
        \midrule
        \multicolumn{4}{@{}l}{\textbf{BlockStackKitchen}} \\
        1 & Reach green cube & End effector within $0.2\,\mathrm{m}$ & --- \\
        2 & Reach wood cube & End effector within $0.2\,\mathrm{m}$ & --- \\
        3 & Lift green cube & Cube height $>0.09\,\mathrm{m}$ & 1 \\
        4 & Lift wood cube & Cube height $>0.09\,\mathrm{m}$ & 2 \\
        5 & Green cube on tray & At least 80\% overlap with tray; gripper open & 3 \\
        6 & Wood cube on tray & At least 80\% overlap with tray; gripper open & 4 \\
        7 & Green cube on wood cube & At least 50\% overlap with wood cube; gripper open & 5, 6 \\
        \bottomrule
    \end{tabular}
\end{table}

\FloatBarrier
%\clearpage
\section{Additional Open-Loop Driving Results}
\label{app:additional_driving_results}

In the main report, we utilized ADE as our metric; here, we report minADE.

\subsection{Minimum ADE}
\label{app:minade}

\begin{figure}[H]
    \centering
    \includegraphics[width=\linewidth]{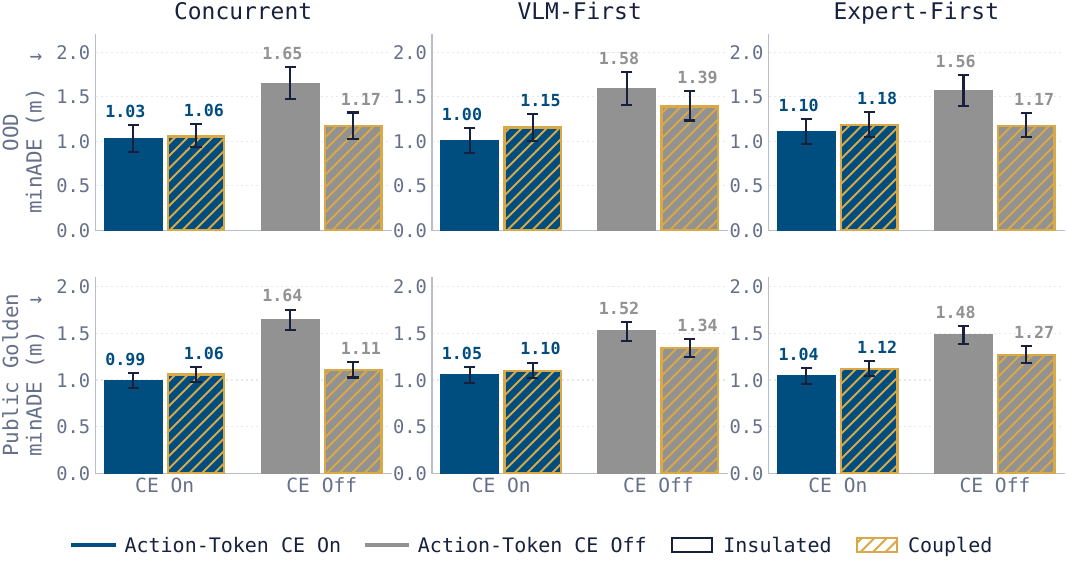}
    \caption{\textbf{Open-loop driving across training histories.} Selected-trajectory minADE on the Public Golden dataset and OOD evaluation dataset. Bars show condition means, and error bars show 95\% clip-cluster bootstrap confidence intervals for those means. Lower is better.}
    \label{fig:concurrent_driving_endpoints_minADE}
\end{figure}

Minimum ADE uses the ground truth to choose the best of six sampled trajectories. At the 300k endpoints, its routing pattern is consistent with the selected-trajectory ADE results: removing action-token supervision is more costly under insulation than under coupling across the concurrent, VLM-first, and expert-first histories on both evaluation datasets. Concurrent training achieves the lowest observed minADE on Public Golden, whereas VLM-first training achieves the lowest on OOD (Figure~\ref{fig:concurrent_driving_endpoints_minADE}). 

\FloatBarrier
\section{Additional Closed-Loop Driving Results}
\label{app:closed_loop_driving_results}

\begin{figure}[H]
    \centering
    \includegraphics[width=1\linewidth]{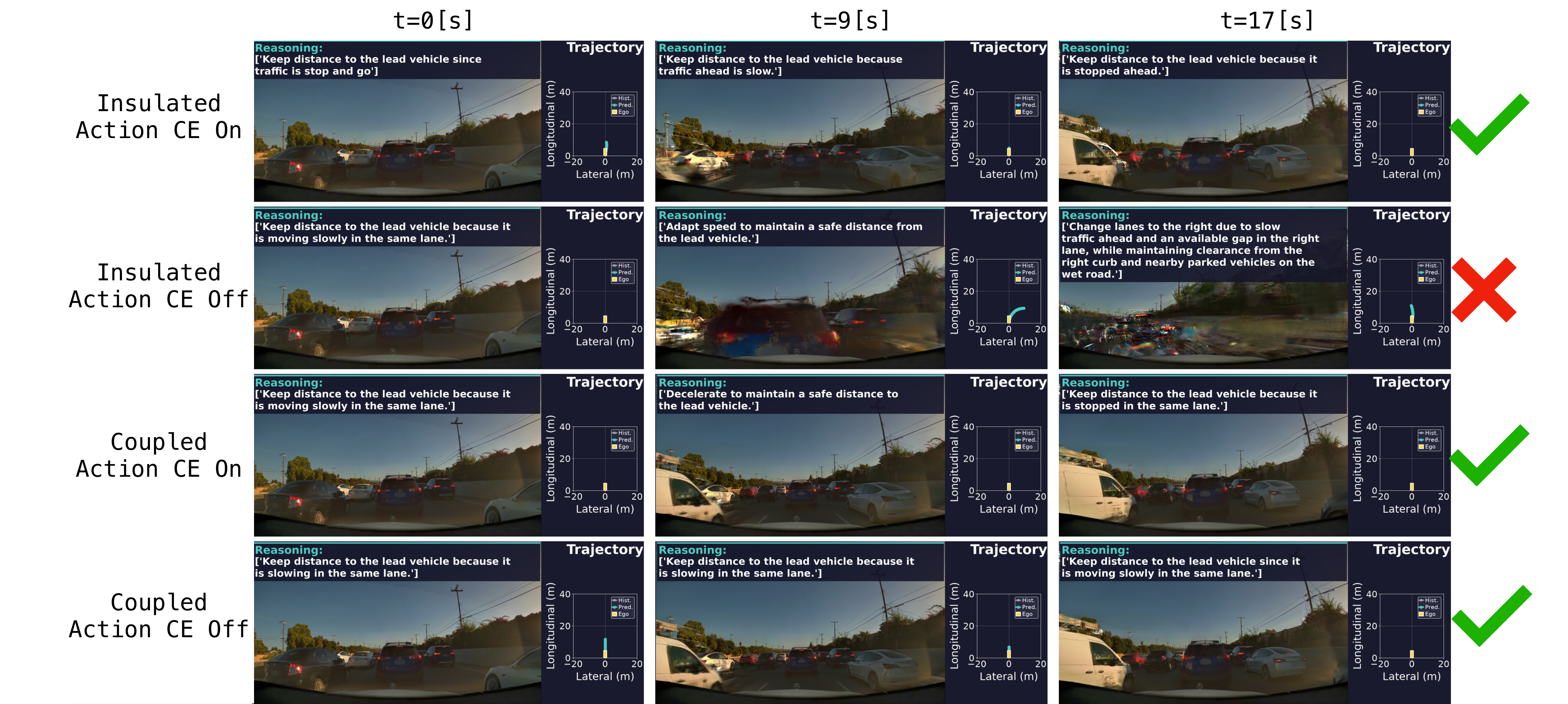}
    \caption{\textbf{Concurrent-training AlpaSim case study.} Four conditions in the same traffic scene. The insulated rollout without action-token supervision has an at-fault collision. The displayed rollouts for the other three conditions succeed. This example illustrates a failure mode behind the aggregate results in Table~\ref{tab:alpasim_concurrent_300k}.}
    \label{fig:alpasim_concurrent_qualitative}
\end{figure}

Tables~\ref{tab:alpasim_concurrent_300k}, \ref{tab:alpasim_expert_first_300k}, and~\ref{tab:alpasim_vlm_first_300k} report the full AlpaSim results. All conditions use the same 827 scenes, with five stochastic rollouts averaged within each scene. AlpaSim terminates or truncates after safety events and censors trajectories at 4\,m from the ground-truth trajectory. Arrows indicate the preferred direction, and the best value in each row is bolded. We moreover provide a qualitative demonstration of policy performance in Figure~\ref{fig:alpasim_concurrent_qualitative}.

\begin{minipage}{\linewidth}
    \centering
    \small
    \setlength{\tabcolsep}{5pt}
    \renewcommand{\arraystretch}{1.05}
    \captionsetup{hypcap=false}\captionof{table}{\textbf{Concurrent training: AlpaSim metrics.}}
    \label{tab:alpasim_concurrent_300k}
    \begin{tabular*}{\linewidth}{@{\extracolsep{\fill}}lrrrr@{}}
        \toprule
        & \multicolumn{2}{c}{\textbf{Insulated}} & \multicolumn{2}{c}{\textbf{Coupled}} \\
        \cmidrule(lr){2-3}\cmidrule(l){4-5}
        \textbf{Metric} & \textbf{Tokens on} & \textbf{Tokens off} & \textbf{Tokens on} & \textbf{Tokens off} \\
        \midrule
        Progress $\uparrow$ & 0.600 & 0.425 & \textbf{0.703} & 0.601 \\
        Distance traveled (m) $\uparrow$ & 116.68 & 78.63 & \textbf{139.59} & 114.78 \\
        Duration fraction $\uparrow$ & 0.590 & 0.363 & \textbf{0.634} & 0.541 \\
        \addlinespace[4pt]
        Collision, any $\downarrow$ & 0.225 & 0.311 & \textbf{0.164} & 0.211 \\
        Collision, at fault $\downarrow$ & \textbf{0.087} & 0.187 & 0.090 & 0.127 \\
        Off-road $\downarrow$ & 0.161 & 0.304 & \textbf{0.136} & 0.184 \\
        Off-road or at-fault collision $\downarrow$ & 0.248 & 0.491 & \textbf{0.226} & 0.311 \\
        \addlinespace[4pt]
        Distance to ground truth (m) $\downarrow$ & \textbf{2.378} & 2.471 & 2.700 & 2.732 \\
        Plan deviation $\downarrow$ & 0.380 & 0.508 & \textbf{0.321} & 0.378 \\
        Wrong lane $\downarrow$ & 0.201 & \textbf{0.170} & 0.230 & 0.206 \\
        \bottomrule
    \end{tabular*}
\end{minipage}

Coupling increases route progress without action-token supervision in both staged histories: from 0.538 to 0.604 after expert-first training and from 0.555 to 0.587 after VLM-first training. With action-token supervision, coupling increases progress from 0.597 to 0.625 after expert-first training, while the VLM-first values are nearly identical (0.648 insulated and 0.646 coupled). The highest observed route progress across all three histories is 0.703 under concurrent coupled training with action-token supervision (Table~\ref{tab:alpasim_concurrent_300k}).

\begin{minipage}{\linewidth}
    \centering
    \small
    \setlength{\tabcolsep}{5pt}
    \renewcommand{\arraystretch}{1.05}
    \captionsetup{hypcap=false}\captionof{table}{\textbf{Expert-first training: AlpaSim metrics.}}
    \label{tab:alpasim_expert_first_300k}
    \begin{tabular*}{\linewidth}{@{\extracolsep{\fill}}lrrrr@{}}
        \toprule
        & \multicolumn{2}{c}{\textbf{Insulated}} & \multicolumn{2}{c}{\textbf{Coupled}} \\
        \cmidrule(lr){2-3}\cmidrule(l){4-5}
        \textbf{Metric} & \textbf{Tokens on} & \textbf{Tokens off} & \textbf{Tokens on} & \textbf{Tokens off} \\
        \midrule
        Progress $\uparrow$ & 0.597 & 0.538 & \textbf{0.625} & 0.604 \\
        Distance traveled (m) $\uparrow$ & 115.25 & 104.28 & \textbf{119.83} & 116.13 \\
        Duration fraction $\uparrow$ & \textbf{0.599} & 0.461 & 0.561 & 0.553 \\
        \addlinespace[4pt]
        Collision, any $\downarrow$ & 0.230 & 0.305 & \textbf{0.165} & 0.232 \\
        Collision, at fault $\downarrow$ & \textbf{0.061} & 0.182 & 0.097 & 0.139 \\
        Off-road $\downarrow$ & 0.159 & 0.247 & \textbf{0.134} & 0.182 \\
        Off-road or at-fault collision $\downarrow$ & \textbf{0.220} & 0.429 & 0.231 & 0.321 \\
        \addlinespace[4pt]
        Distance to ground truth (m) $\downarrow$ & \textbf{2.152} & 2.569 & 2.850 & 2.514 \\
        Plan deviation $\downarrow$ & \textbf{0.346} & 0.549 & 0.353 & 0.413 \\
        Wrong lane $\downarrow$ & \textbf{0.190} & 0.202 & 0.288 & 0.222 \\
        \bottomrule
    \end{tabular*}
\end{minipage}

\medskip

\begin{minipage}{\linewidth}
    \centering
    \small
    \setlength{\tabcolsep}{5pt}
    \renewcommand{\arraystretch}{1.05}
    \captionsetup{hypcap=false}\captionof{table}{\textbf{VLM-first training: AlpaSim metrics.}}
    \label{tab:alpasim_vlm_first_300k}
    \begin{tabular*}{\linewidth}{@{\extracolsep{\fill}}lrrrr@{}}
        \toprule
        & \multicolumn{2}{c}{\textbf{Insulated}} & \multicolumn{2}{c}{\textbf{Coupled}} \\
        \cmidrule(lr){2-3}\cmidrule(l){4-5}
        \textbf{Metric} & \textbf{Tokens on} & \textbf{Tokens off} & \textbf{Tokens on} & \textbf{Tokens off} \\
        \midrule
        Progress $\uparrow$ & \textbf{0.648} & 0.555 & 0.646 & 0.587 \\
        Distance traveled (m) $\uparrow$ & 128.59 & 104.96 & \textbf{129.10} & 113.02 \\
        Duration fraction $\uparrow$ & \textbf{0.624} & 0.508 & 0.588 & 0.512 \\
        \addlinespace[4pt]
        Collision, any $\downarrow$ & 0.208 & 0.264 & 0.204 & \textbf{0.175} \\
        Collision, at fault $\downarrow$ & \textbf{0.076} & 0.141 & 0.117 & 0.106 \\
        Off-road $\downarrow$ & \textbf{0.126} & 0.215 & 0.166 & 0.199 \\
        Off-road or at-fault collision $\downarrow$ & \textbf{0.203} & 0.355 & 0.282 & 0.305 \\
        \addlinespace[4pt]
        Distance to ground truth (m) $\downarrow$ & 2.515 & \textbf{2.512} & 2.675 & 2.829 \\
        Plan deviation $\downarrow$ & \textbf{0.360} & 0.442 & 0.412 & 0.418 \\
        Wrong lane $\downarrow$ & \textbf{0.228} & 0.244 & 0.238 & 0.237 \\
        \bottomrule
    \end{tabular*}
\end{minipage}

\FloatBarrier

\end{document}